\documentclass[10pt,twocolumn,letterpaper]{article}

\usepackage[pagenumbers]{wacv} 

\usepackage{graphicx}
\usepackage{amsmath}
\usepackage{amssymb}
\usepackage{booktabs}

\usepackage{color}
\usepackage{xspace}
\usepackage{xcolor}
\usepackage{multirow}

\definecolor{wacvblue}{rgb}{0.21,0.49,0.74}
\usepackage[pagebackref,breaklinks,colorlinks,allcolors=wacvblue]{hyperref}

\usepackage[capitalize]{cleveref}
\crefname{section}{Sec.}{Secs.}
\Crefname{section}{Section}{Sections}
\Crefname{table}{Table}{Tables}
\crefname{table}{Tab.}{Tabs.}

\def\wacvPaperID{272} 
\def\confName{WACV}
\def\confYear{2026}

\begin{document}

\title{Optimization-Free Style Transfer for 3D Gaussian Splats}

\author{Raphael Du Sablon and David Hart\\
East Carolina University\\
{\tt\small hartda23@ecu.edu}
}
\maketitle

\begin{abstract}
   The task of style transfer for 3D Gaussian splats has been explored in many previous works, but these require reconstructing or fine-tuning the splat while incorporating style information or optimizing a feature extraction network on the splat representation. We propose a reconstruction- and optimization-free approach to stylizing 3D Gaussian splats, allowing for direct stylization on a .ply or .splat file without requiring the original camera views. This is done by generating a graph structure across the implicit surface of the splat representation. A feed-forward, surface-based stylization method is then used and interpolated back to the individual splats in the scene. 
   This also allows for fast stylization of splats with no additional training, achieving speeds under 2 minutes even on CPU-based consumer hardware. We demonstrate the quality results this approach achieves and compare to other 3D Gaussian splat style transfer methods. 
   Code is publicly available at \href{https://github.com/davidmhart/FastSplatStyler}{https://github.com/davidmhart/FastSplatStyler}.
\end{abstract}

\section{Introduction}
\label{sec:intro}

3D Gaussian Splats (3DGS) \cite{kerbl20233d} have quickly gained high amounts of interest in the research community for their ability to capture real world scenes with incredibly high visual fidelity. This is all done while maintaining an intrinsic representation of ``splats'' that are easily described and stored. This representation, however, is not conducive to simple editing and modification techniques that are common to other 3D representations such as 3D meshes.

One such modification includes style transfer, the task of transferring the low-level artistic properties of a style image onto a content representation. While this area has a deep history for image-to-image style transfer \cite{gatys2016image,johnson2016perceptual,li2017universal,li2019learning,deng2022stytr2,styleID}, the process of stylizing onto a 3DGS has only recently been explored. Some works have presented stylization options for 3DGS, but they rely on starting the reconstruction process from scratch while including style features in the process \cite{jain2024stylesplat,galerne2024sgsst}. Some methods have aimed to increase the stylization speed by not requiring reconstruction, using post processing techniques instead. G-style \cite{kovacs2024} performs a special fine-tuning on an existing 3DGS to increase convergence speed. StyleGaussian \cite{liu2024stylegaussian} can operate on an existing 3DGS, but it does require learning a content feature extraction network on the 3DGS before the stylization is possible. In short, none of the current works are able to perform style transfer on an existing 3DGS without reconstructing or reoptimizing the 3DGS.

\begin{figure}
\begin{center}
       \begin{tabular}{cc}       
        \includegraphics[width=0.42\linewidth]{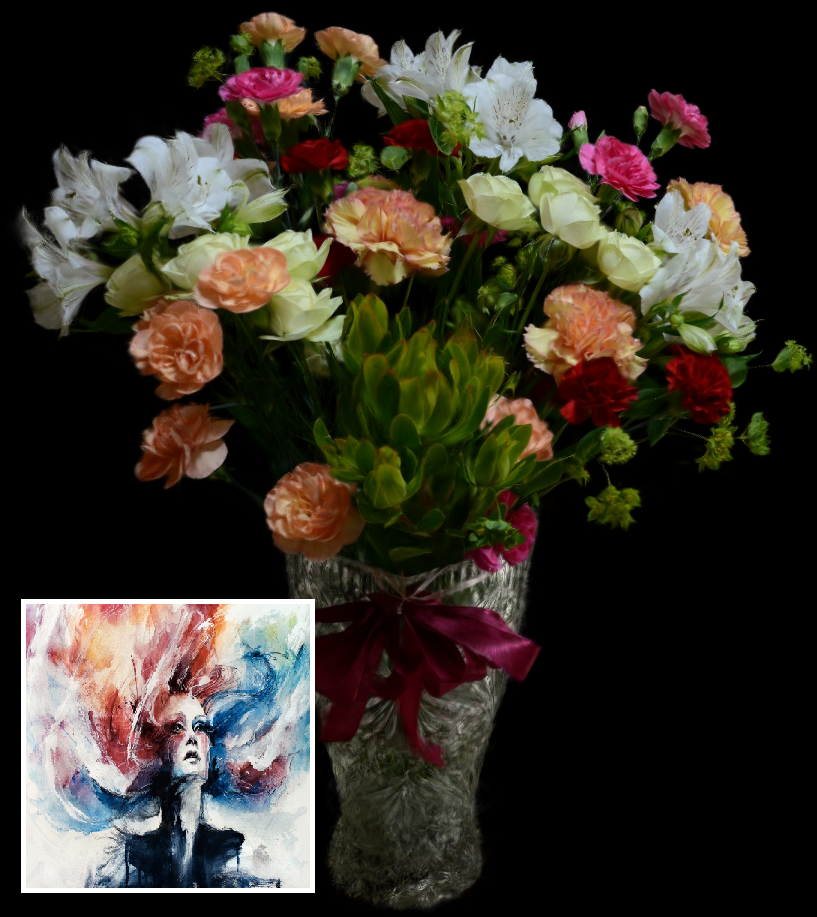}
        &
        \includegraphics[width=0.42\linewidth]{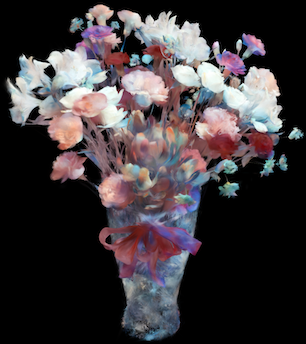}
        \\
        Style and 3DGS
        &
        Stylized 3DGS
        
    \end{tabular}
\end{center}
   \caption[Style Transfer Results, Flower Bouquet]{A 3D Gaussian splat with geometrically complex content and an image containing a distinct artistic style (Left) and a 3D Gaussian splat of the same content after style transfer using our approach (Right).}
\label{fig:introresults}
\end{figure}

In this work, we present a new approach for style transfer for 3DGS that does not require reconstructing the splat, finetuning a network, or any other form of optimization. It does so by building a graph structure across the implicit surface of the 3DGS. This graph is constructed in such a way as to allow standard image convolution networks to be applied to the surface \cite{hart2023interpolated}. Then, a standard pretrained image style transfer network can be used \cite{li2019learning} and the colors on the graph can be interpolated back to the original splats contained within the 3DGS scene. An example result is given in Fig.~\ref{fig:introresults}. Our approach was designed while considering many possible variations and a thorough ablation study on the algorithm is provided in this work.

Our approach has some immediately apparent advantages. First, our approach does not require access to the original images from which the splat was constructed. This means that publicly available \textit{.splat} or \textit{.ply} files from public repositories or outputted from apps such as Scaniverse~\cite{Scaniverse} can be stylized with our method. Second, since the style colors are stored with the individual splats in a scene, the stylization can be saved to the splat file and is not viewer dependent. Third, our approach is significantly faster than approaches that require reconstructing a 3DGS from scratch. Our approach is also primarily CPU-based, allowing it to run quickly even on standard consumer-grade hardware, with stylization times under 2 minutes for most 3DGS sizes. Additionally, we will show that our approach maintains high visual quality, especially for single-object splats, and has comparable quality to other state-of-the-art approaches for larger scenes.

In summary are contributions are as follows:

\begin{itemize}
    \item A novel surface-based approach for stylizing 3DGS scenes that can run on a CPU without additional optimization or training.
    \item An ablation on techniques to improve the quality of results.
    \item Qualitative and speed comparisons to state-of-the-art 3DGS stylization approaches.
\end{itemize}

\section{Related Work}

\subsection{Neural Radiance Fields and Gaussian Splats}

In the various approaches to 3D rendering, the development of effective neural networks rapidly led to new iterations of and improvements on classical methods
\cite{kovacs2024surface,schwarz2022voxgraf, xu2024instantmesh}. In particular, the field saw a revolution with the introduction of NeRFs, Neural Radiance Fields, introduced by \cite{mildenhall2021nerf}. 
NeRF is able to represent geometrically complex scenes involving difficult optical phenomena as five-dimensional radiance fields.
A key insight of NeRF is to use a neural network as a mechanism for storing the learned visual information of a scene.
The differentiability of the rendering process allows for optimization to be performed and for a loss function to be defined on the final 2D view of the representation.
A particular feature of this algorithm is that it allows for rendering new views of 3D scenes, a process termed \textit{novel view synthesis}. 
Decades-old novel view synthesis techniques do exist \cite{chen2023view}, but as with other areas, the field saw a marked improvement with the development of modern machine learning techniques \cite{hedman2018deep, sitzmann2019deepvoxels, sitzmann2019scene, xu2024instantmesh}.


3D Gaussian Splatting for Real-Time Radiance Field Rendering \cite{kerbl20233d} builds on the success of radiance fields while 
eliminating the use of a neural network at render time. In doing so, Gaussian splatting enabled real-time rendering of novel-view synthesis at high resolutions and qualities with a simply described representation. \textit{3D Gaussian Splats} (or simply ``splats") are a point-cloud-like primitive that is differentiable, unstructured, and easily projected to 2D for rendering. An optimization algorithm for fine-tuning the properties of Gaussian splats and a novel rasterizer allows for back-propagation while also being fast to render high quality scenes. Alpha-blending of the Gaussians as they are projected to 2D ensures good textural detail and smooth geometries, at least in regions of high splat density \cite{hu2024low}.

This unique approach is of high interest in the research community because of its incredible visual quality for novel view synthesis of real scenes. Many works have followed that explore improving the reconstruction process and representation \cite{Yu_2024_CVPR, Fu_2024_CVPR, Huang2024gaussian2D}. Some works have also explored pruning and compressing 3D Gaussian Splats \cite{niedermayr2024compress1, fan2023compress2, navaneet2023compress3, lee2024compress4, morgenstern2023compress5} or even editing them \cite{jaganathan2024ice,igs2gs,gsedit}, but the ability to modify such representations is still very limited given that the location of individual splats do not necessarily correlate with true geometry or semantic regions of an object.


\subsection{Style Transfer}

While some classical approaches exist \cite{painterlyrendering}, modern image-to-image style transfer begins with the foundational work of Gatys \etal \cite{gatys2016image}. This original approach treated style transfer as an optimization problem, where the content image was modified until it optimally matched a style image at the features extracted from a VGG network \cite{simonyan2014very}. In the decade since, many new approaches have emerged for performing style transfer. These include those that improved on the original optimization approach \cite{gatys2017controlling}, feed-forward techniques \cite{johnson2016perceptual}, autoencoder-based reconstruction methods \cite{li2017universal,li2019learning}, vision transformer models \cite{wu2021styleformer, deng2022stytr2}, and, most recently, diffusion-based setups \cite{Zhang_2023_inst,styleID}.

In addition to style transfer for images, other works have applied style in other contexts, such as for video \cite{huang2017, ruder2018artistic, ebsynth, Ye_2025_CVPR}, 360$^\circ$ images \cite{ruder2018artistic,hart2023interpolated}, multi-view setups \cite{lightfield1,lightfield2}, 3D meshes \cite{stylit,styblit,hart2023interpolated}, and even fluid simulation \cite{fluidstyle}.
Specifically, our work builds on SelectionConv \cite{hart2022selectionconv,hart2023interpolated}, which uses a graph to apply style to 3D meshes and other nonstandard domains.

\begin{figure*}[t!]
    \centering
    \includegraphics[width=1.00\linewidth]{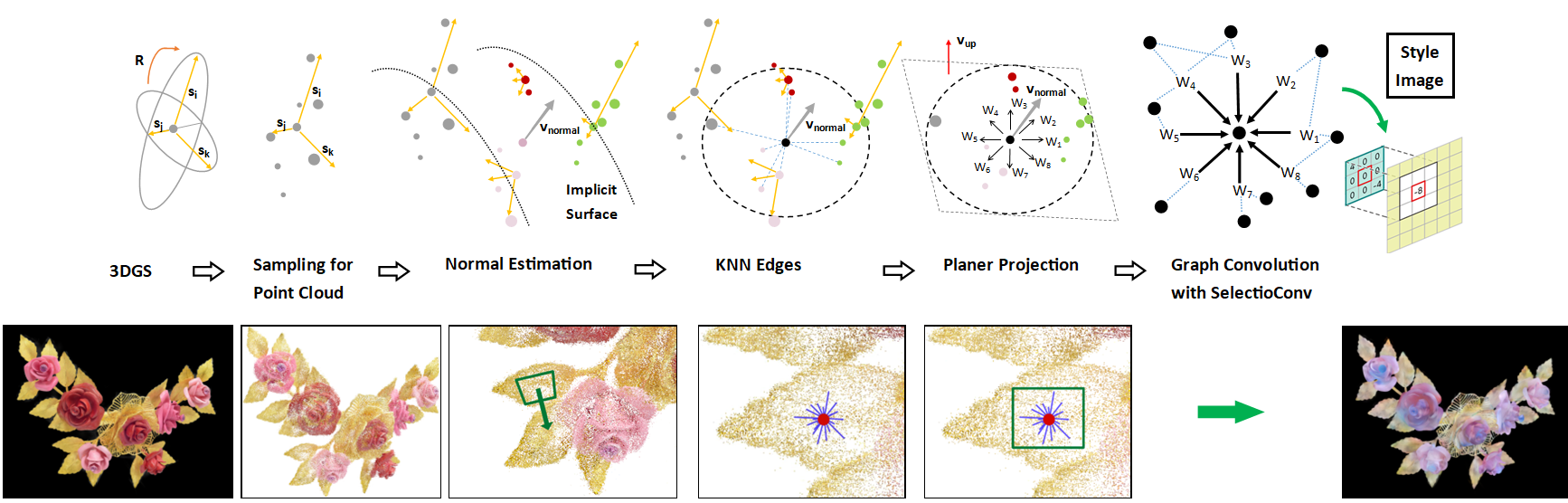}
    \caption[Convolution on Gaussian splats]{Overview of our approach. A graph construction pipeline takes an existing Gaussian Splat scence and samples it as a point cloud. Normal vectors are approximated and points in the point cloud are connected using a K-Nearest-Neighbors approach. A local planar approximation of the graph at each point in the graph allows for convolution on the splat using 2D image CNN weights. A style transfer network is then converted into the graph space using SelectionConv \cite{hart2023interpolated}. The resulting graph is interpolated back to existing splat center to modify the color values, generating the final stylized result. The pipeline is shown in both a technical version (Top) and full splat visualization (Bottom).}
    \label{fig:Process}
\end{figure*}

\subsection{Style Transfer for 3D Gaussian Splats}

There has been considerable attention paid to attempting to translate the success of image-to-image style transfer to the 3D realm of radiance fields and Gaussian splatting. Zhang \etal \cite{zhang2022arf} presented an early method for applying style transfer techniques to 3D radiance fields, with others works to follow \cite{chiang2022stylizing,liu2023stylerf}. 
For Gaussian splats, most approaches focus on stylizing the splats as the scene is being reconstructed.
The work of Sahora \etal \cite{saroha2024gaussian} builds on the AdaIN layer \cite{huang2017arbitrary} for aligning content and style features during reconstruction.
VGG features are used by Zhang \etal \cite{zhang2024stylizedgs}, which constructs the loss function in a way that evolves the style spatially along the objects represented in the 3D Gaussian. Jain \etal \cite{jain2024stylesplat} leverage segmentation models to allow for style transfers to select portions the 3D Gaussian that are being styled. A different niche is filled by Galerne \etal \cite{galerne2024sgsst}, which applied a new method of addressing scale to tackle style transfers to ultra-high resolution 3D Gaussians. 

In comparison to the works mentioned, some approaches have attempted to stylize without requiring reconstructing the scene from scratch. The method of Kovacs \etal \cite{kovacs2024} can fine-tune an already existing reconstruction to incorporate style features. Liu \etal \cite{liu2024stylegaussian} used a non-reconstruction-based approach and is most comparable to this work. They introduced a KNN 3D CNN model to reduce operations that break multi-view consistency and it employs a unique feature extraction network that associates VGG features with individual splats that is view-independent and does not require reconstruction. This approach, however, is still not optimization-free since it requires training the feature extraction network and potentially the stylization network. In comparison, our graph construction approach is well defined and can be applied to any Gaussian splat scene with no additional fine tuning or training. It can also be applied to a splat even if the original camera views used for reconstruction cannot be obtained.



\section{Methodology}

The proposed method of stylizing a 3D Guassian Splat (3DGS) builds on the surface-based CNN called SelectionConv \cite{hart2022selectionconv,hart2023interpolated} and the style transfer network of Li \etal \cite{li2019learning} for its foundational building block. This relies on constructing a properly oriented graph along the surface. An overview of our approach is show in Fig.~\ref{fig:Process}. The graph construction process, along with various improvements designed for 3DGS representations are described in the following sections.

\subsection{Graph Construction}

Interpolated SelectionConv is a framework for convolutional neural network operations that can operate on meshes or surfaces \cite{hart2023interpolated}. It allows for neural networks originally configured for and trained on 2D images to operate on complex 3D meshes by representing the mesh as a graph network. Points sampled along the surface of the mesh become the nodes in the graph and edges are connected based on proximity. The graph is then fed into a modified network that mimics the operations and orientation of standard image convolution. Interpolation then handles the variations in the relations between nodes on the mesh or surface. 

In applying this approach to a 3DGS, a key difference is immediately apparent: meshes have points that are fixed to the surface while splats can be scattered volumetrically throughout the object. Although this is the case, we have found that 3DGS representations tend to place most splats near the surface of the object. This creates a pseudo implicit surface that does not vary drastically in depth and is sufficient for the process of style transfer.

Moving forward with this assumption, we take the following steps to construct an oriented graph.

\paragraph{Point Cloud Sampling:} By default, the centers of each individual splat are treated as points in a point cloud. These become the nodes in the graph. If additional points are desired, they can be sampled from the 3D Gaussian distributions. Undesired points can be removed by filtering. More details are provided in Secs.~\ref{sec:sampling} and \ref{sec:filtering}.

\paragraph{Normal Estimation:} The graph network employed by Interpolated SelectionConv requires up vectors and accurate normal vectors in order to perform the interpolation and weight assignment. This allows the orientation of the cardinal and ordinal directions to remain consistent for the convolution operation on both the 3D surface and on an ordered 2D pixel grid. For the normal vectors, we find that standard point cloud normal-estimation methods are sufficient. In this work, the Ball-Pivoting sampling method of \cite{bernardini2002ball} is used via Open3D's implementation \cite{Open3D}. 

\paragraph{KNN Edges:} Edges for the graph are generated using the common K-Nearest-Neighbors approach. This connects the points in the sampled point cloud to each other based on proximity. For this work, we use a K-value of 16 when constructing the graphs.

\paragraph{Selection through Planar Projection:} To determine the oriented direction for each edge, a local planar projection is created using the previously defined normal and an up vector. An arbitrary up vector can be used or potentially one can be derived from geometric information implied by the Gaussian splat if desired. Graham-Schimdt orthogonalization between the up vector and normal vector create a locally planar axes for each node in the graph. This allows defining the local oriented direction of each edge relative to each point. Completing this final step for each node provides a graph with all needed information for performing the stylization.
 




\subsection{Stylization}

Once the graph is properly constructed, stylization is straightforward. The Interpolated SelectionConv \cite{hart2023interpolated} graph network copies the weights of the style transfer network of Li \etal \cite{li2019learning} and applies the surface-based stylization to the graph. The style network of \cite{li2019learning} was chosen for its ability to be mimicked by graph convolutions. To clarify the stylization process, we summarize the method and again address the issue of how a surface-based approach is handled in a volumetric graph. 

In \cite{hart2023interpolated}, the convolution operation for each layer was defined over the graph adjacency matrix as

\begin{equation}
    \label{eq:selection-conv}
    \mathbf{X}^{(k+1)} = \sum_m{\tilde{\textbf{S}}_m \mathbf{X}^{(k)} \textbf{W}_m}
\end{equation}
\noindent where $m$ represents a given direction, 
$\tilde{\textbf{S}}_m$ is the adjacency matrix containing edges in that given direction, $\mathbf{X}^{(k)}$ is the current values at each node in the graph, and $\textbf{W}_m$ is the copied learned weights from a standard image CNN, and $\mathbf{X}^{(k+1)}$ is the output values at each node in the graph. To determine which $\tilde{\textbf{S}}_m$ each edge will be a part of, the normal vector at each node and the global up vector are aligned with the surface using Graham-Schimdt orthogonalization. This creates a local axis at each node. Simple dot products are then used to determine which local direction each edge most closely aligns with.

In this original formulation, the assumption is made that, after the axis alignment to the surface, edges are approximately locally planar and depth component of the edge is negligible. This assumption holds well for most high resolution meshes and surfaces. For Gaussian splats, such an assumption can not hold as strictly since points are not restricted to the surface. The 3D variation of edge values for points in a 3DGS is much more drastic than for a standard mesh. 
Even with such variation, however, we empirically find that the locally planar assumption is sufficient and the stylization still works well. The implicit surface that naturally occurs during the creation of a 3DGS places points in a way that meets the original assumptions of the Interpolated SelectionConv graph network.

\begin{figure}[t]
\begin{center}
       \begin{tabular}{cc}  
        \includegraphics[width=0.60\linewidth]{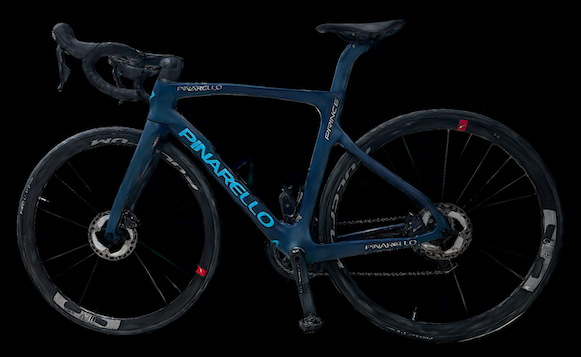}        
        \\
        \includegraphics[width=0.60\linewidth]{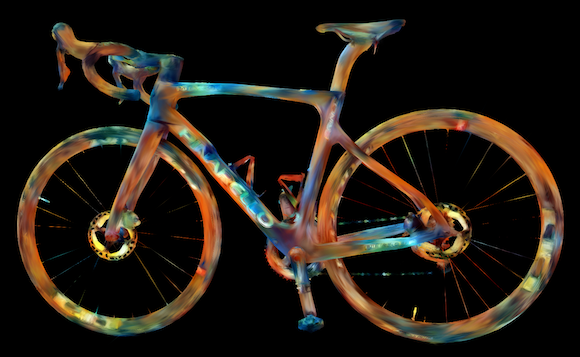}
        &
        \includegraphics[width=0.25\linewidth]{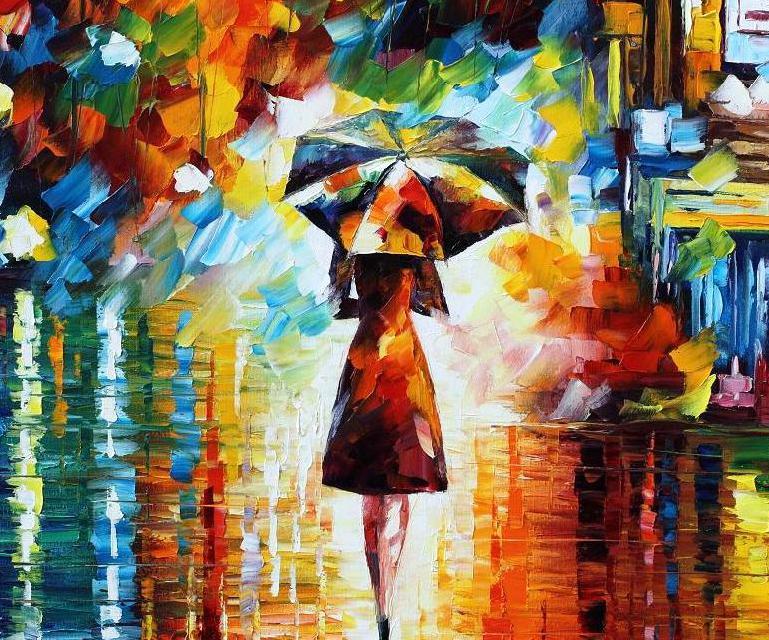}
        
        \\
        \includegraphics[width=0.60\linewidth]{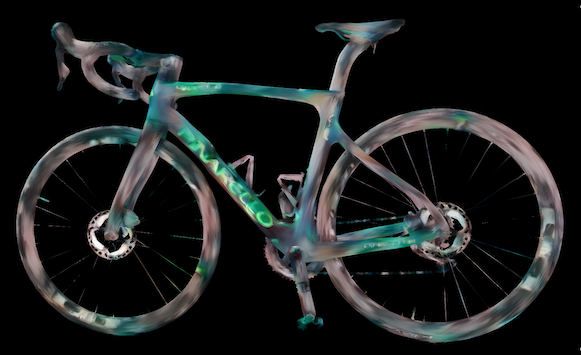}
        &
        \includegraphics[width=0.25\linewidth]{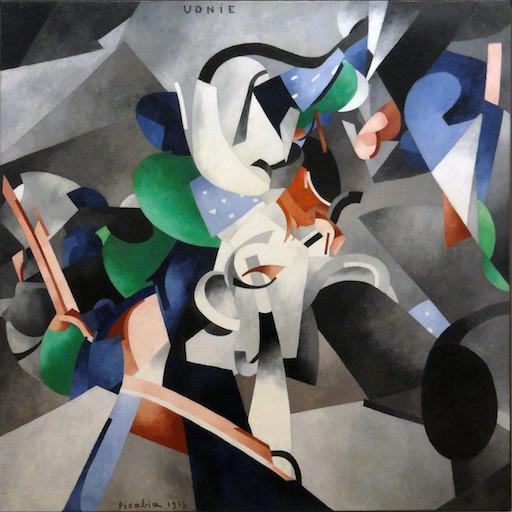}
    \end{tabular}
\end{center}
   \caption[Two Style Transfers on Bicycle Representation]{A 3D Gaussian splat under two different style transfers using our approach.}
\label{fig:bike}
\end{figure}

After the stylization graph network runs, the outputs are the new color values for each node in the graph. These values become the base color values for the individual splats in the 3DGS representation and the stylization is complete. An example of the final result is shown in Fig.~\ref{fig:bike}.

\subsection{Controlling Resolution through Sampling}\label{sec:sampling}

In image style transfer, the resolution of the content image compared to the style image can affect the quality of the stylization \cite{gatys2017controlling}. For 3DGS scene, however, comparing resolutions is difficult to define given the nature of the unstructured data. Empirically, we find that the using a single point per individual splat is usually insufficient to provide high-quality results. Thus, we generate a higher density point cloud by sampling from each individual splat. 

\begin{figure}
\centering
\begin{tabular}{cc}
        \includegraphics[width=0.45\linewidth]{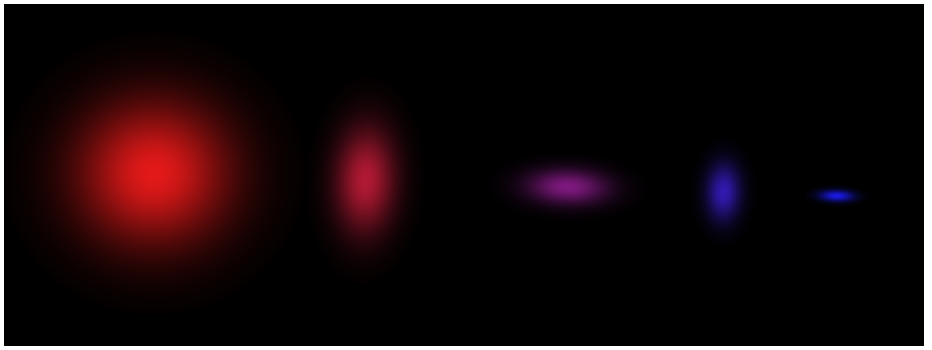}
        &
        \includegraphics[width=0.45\linewidth]{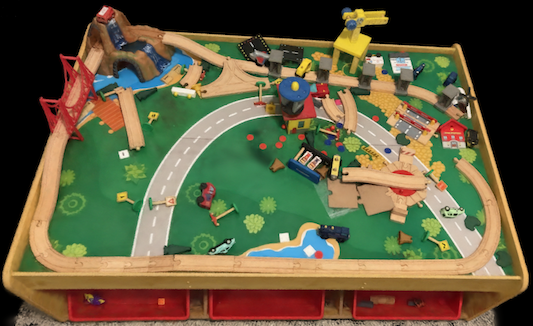}
        \\
        \includegraphics[width=0.45\linewidth]{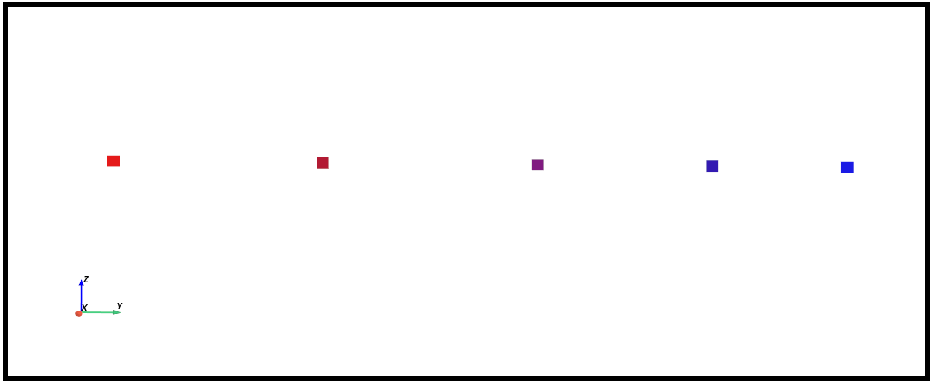}
        &
        \includegraphics[width=0.45\linewidth]{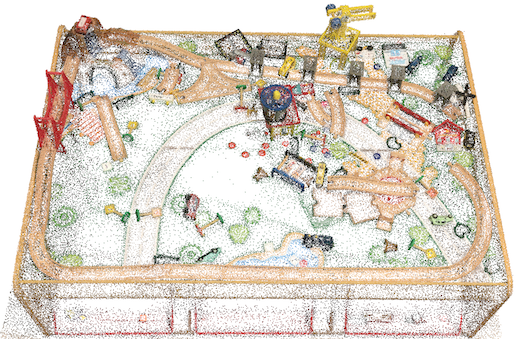}
        \\
        \includegraphics[width=0.45\linewidth]{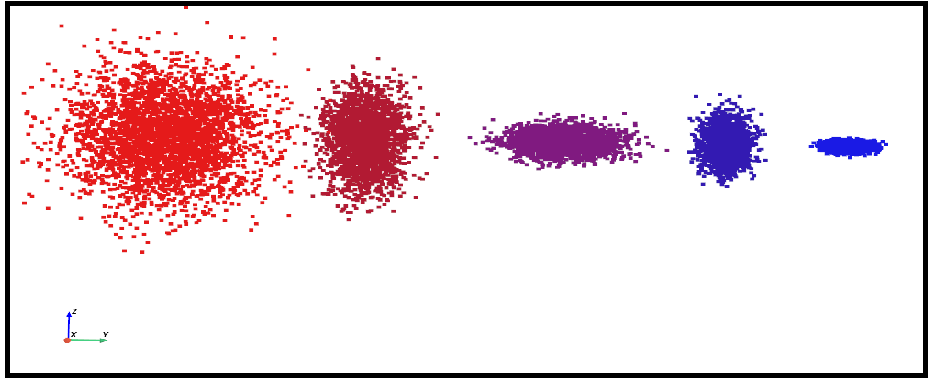}
        &
        \includegraphics[width=0.45\linewidth]{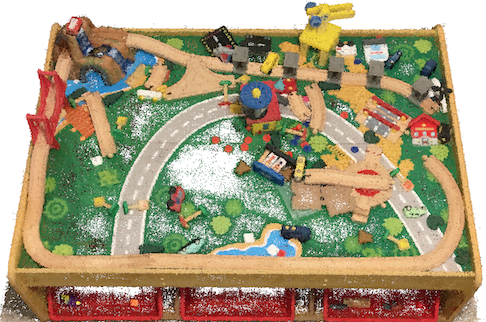}
    \end{tabular}
    
    \caption{A toy example 3D Gaussian splat scene (Left) and high resolution scene of toys (Right). The originally rendered scene (Top) is spatially misrepresented when selecting only the centers of the provided splats (Middle). A higher density point cloud can be generated by extensive sampling within each Gaussian (Bottom). The graph created from this new point cloud provides higher quality results when fed into the style transfer network.
    }
    \label{fig:GaussianSampling5Gs}
\end{figure}

This sampling is intuitive to define since each individual splat is already a 3D normal distribution. If additional sampling is desired, the user specifies the amount of novel points needed. New points are then generated by taking existing splats and sampling points from their 3D distributions. Colors for the points can be assigned to match the Gaussian they are sampled from or can be calculated using a K-Nearest Neighbors average approach.  Splats with larger size and higher opacity are given precedence when sampling new points. This is visualized in Fig.~\ref{fig:GaussianSampling5Gs}. The new points are used in the graph construction process and for the stylization. The stylized results on the higher resolution graph are interpolated back to the original splat centers for the final color assignment. 

\begin{figure}
\centering
    \setlength{\tabcolsep}{1.5pt}
       \begin{tabular}{ccc}       
        \includegraphics[width=0.30\linewidth]{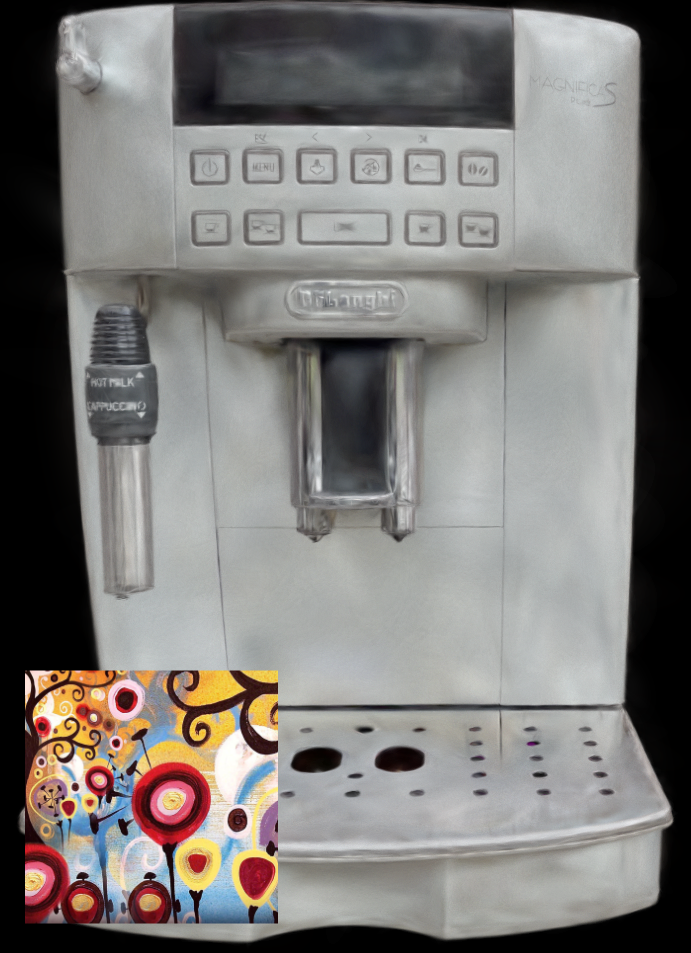}
        &
        \includegraphics[width=0.30\linewidth]{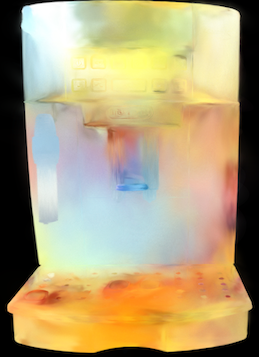}
        &
        \includegraphics[width=0.30\linewidth]{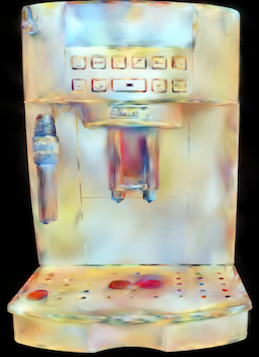}
        \\
        \includegraphics[width=0.30\linewidth]{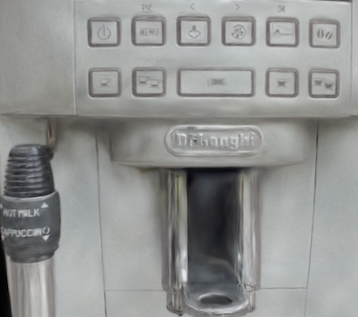}
        &
        \includegraphics[width=0.30\linewidth]{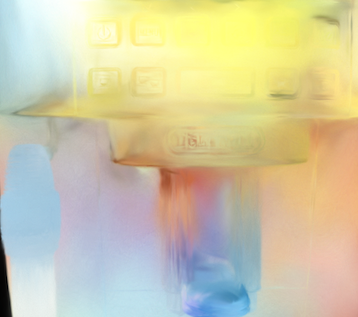}
        &
        \includegraphics[width=0.30\linewidth]{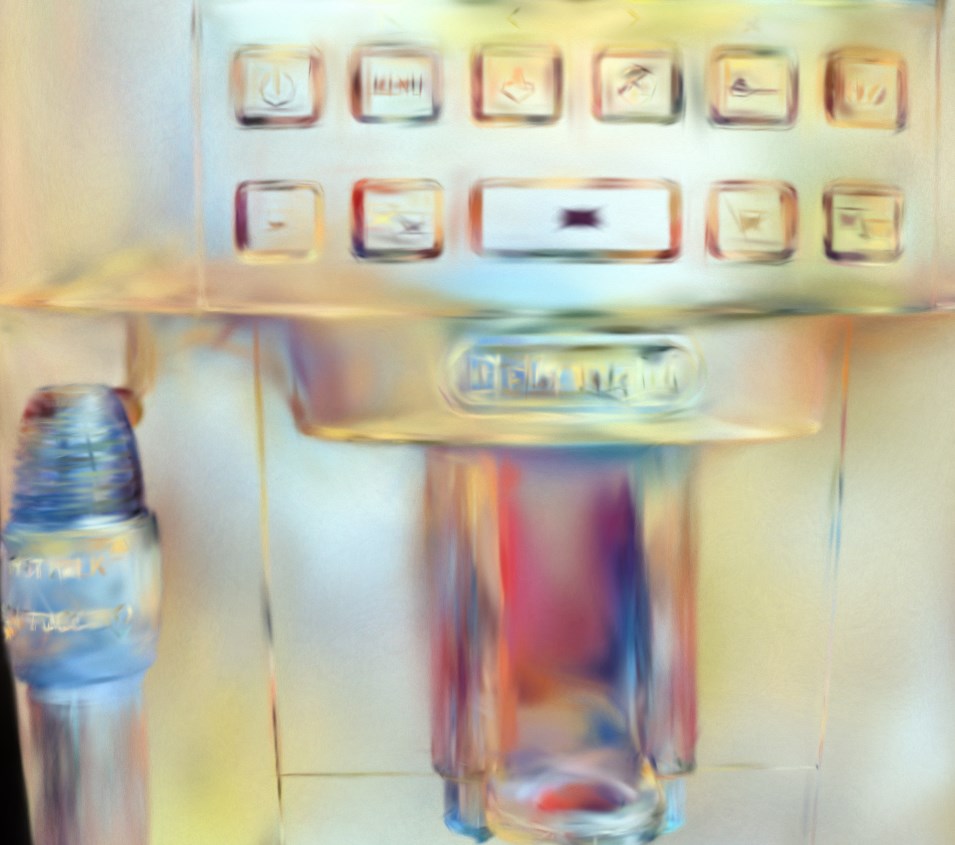}
        \\
        Original
        & Without Sampling
        & With Sampling
        \\
    \end{tabular}
    \caption[Effects of Gaussian Sampling on Style Transfer]{A high quality 3D Gaussian splat (Left) after style transfer using a point cloud created from selecting the centers of individual Gaussians (Middle) and using a point cloud with Gaussian Sampling to reduce information loss (Right) demonstrating the effectiveness of applying Gaussian Sampling. Lettering detail shown in bottom row.}
    \label{fig:coffeesampled}
\end{figure}

We demonstrate the benefit of including our super sampling technique when constructing the point cloud. An example 3DGS stylization, with and without super sampling, is provided in Fig.~\ref{fig:coffeesampled}. The high quality 3DGS shown has considerable intricate details (e.g. legible lettering, indentation around button boundaries, holes in the drip tray). If the super sampling method is not used, the stylization output loses much of the details that were present in the original splat. Additionally, the style transfer itself shows a high degree of blurring and loss in variations. In comparison, when the super sampling method is used, the stylization output preserves many fine details, including lettering and shapes, and has improved variation in colors and contrast.

\subsection{Improving Stylization Quality through Point Filtering}\label{sec:filtering}

Depending on the quality of the 3DGS representation, outlier splats can appear and may throw off the implicit surface. 
This noisy placement of splats comes from poor images, occluded views, or insufficient training time. 
Many publicly available splats from scanning apps tend to have such noisy placement.
To improve the results, we implement a basic filtering algorithm that considers the distance of a point from the average location of its neighbors. Splats that lie a threshold distance from the average of their neighbors were considered outliers and removed from the graph construction process. This improved normal estimation and overall stylization results. More details about the filtering process are provided in the supplemental material.



\section{Results}

\newcommand{\imwidth}{1.00in}
\newcommand{\imheight}{0.60in}

\begin{figure*}[t!]
    \centering
    \begin{tabular}{rccccc}
    & 
    \includegraphics[width=\imwidth,height=\imheight]{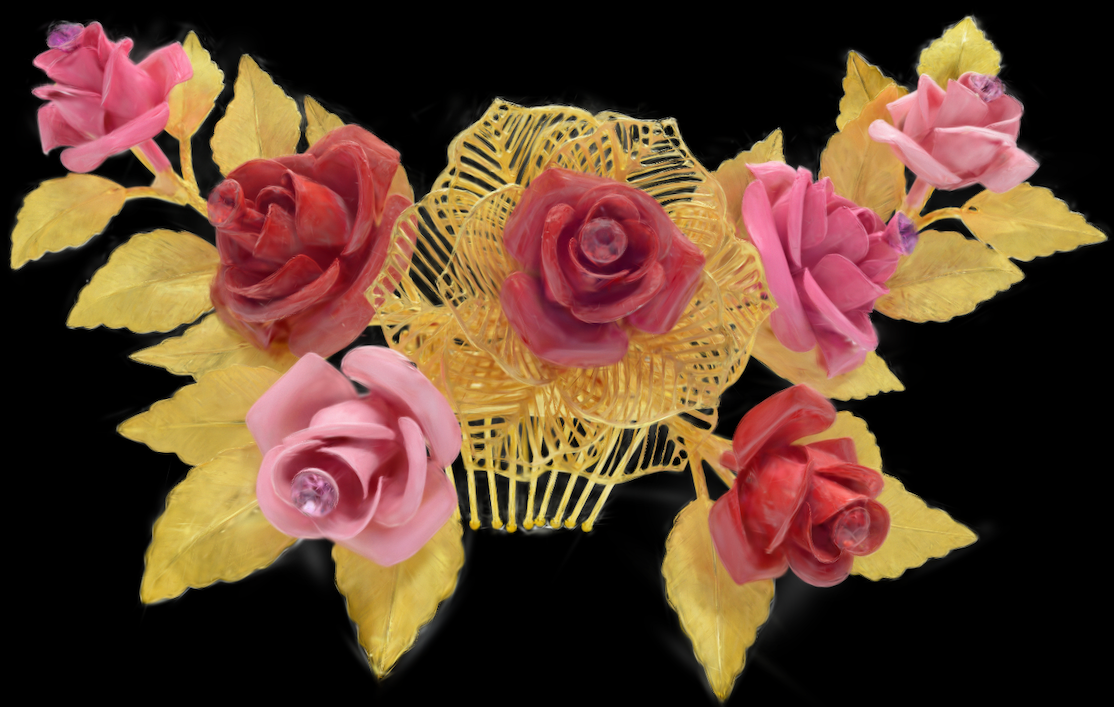} &
    \includegraphics[width=\imwidth,height=\imheight]{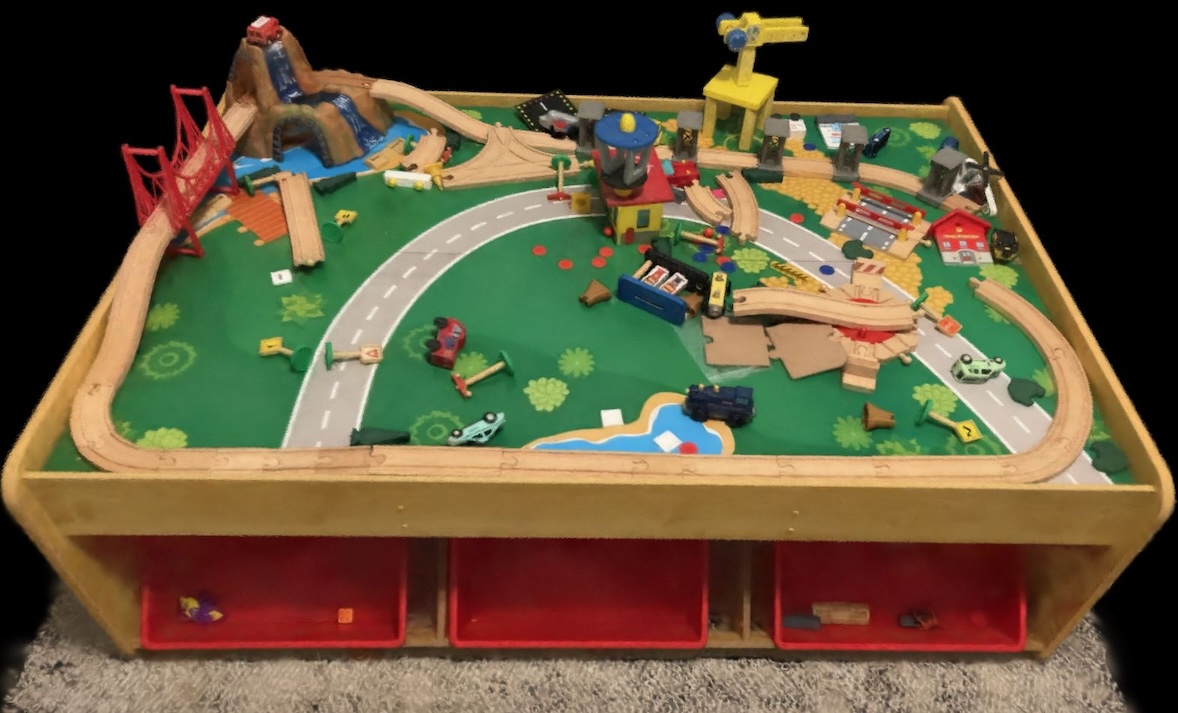}  & 
    \includegraphics[width=\imwidth,height=\imheight]{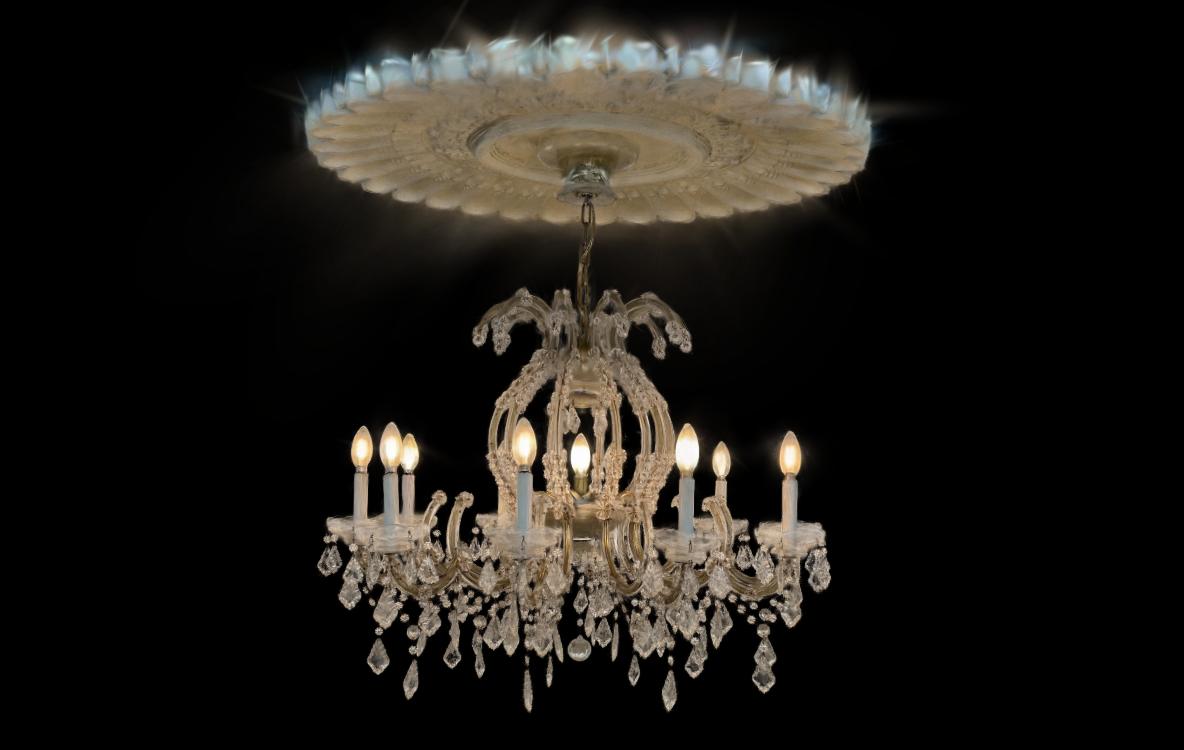} &
    \includegraphics[width=\imwidth,height=\imheight]{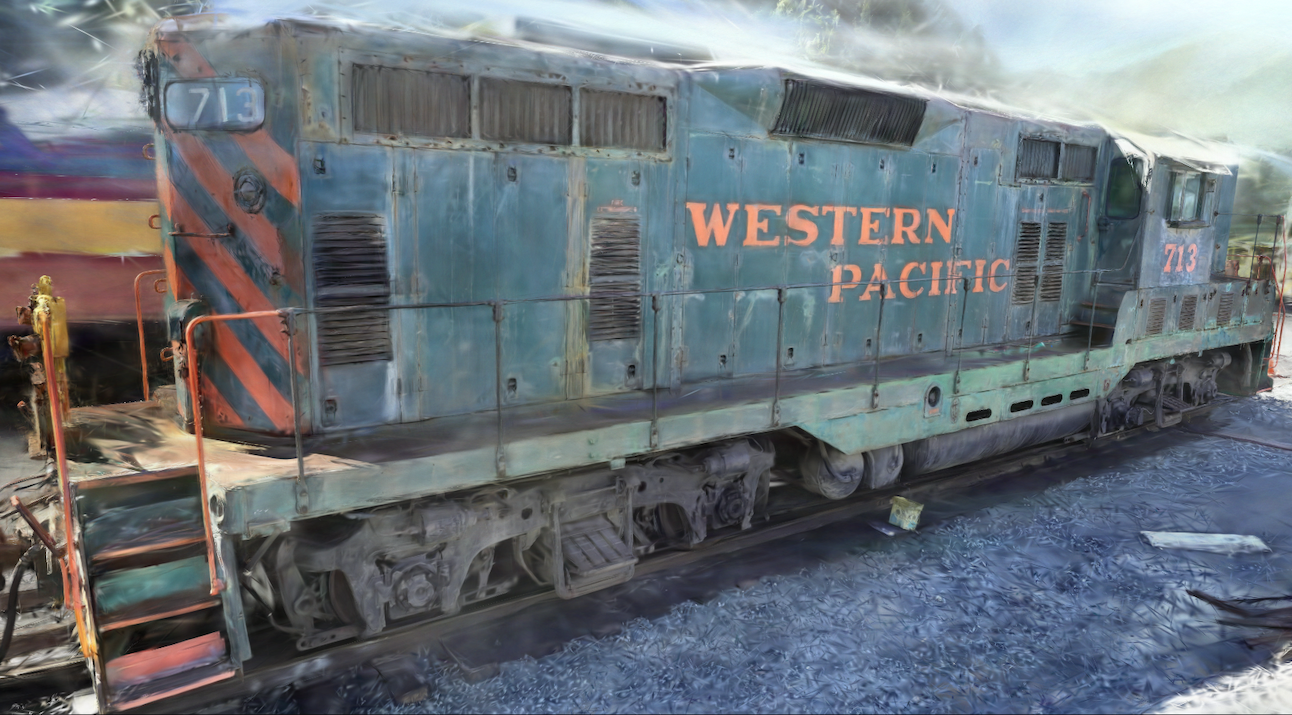} & \includegraphics[width=\imwidth,height=\imheight]{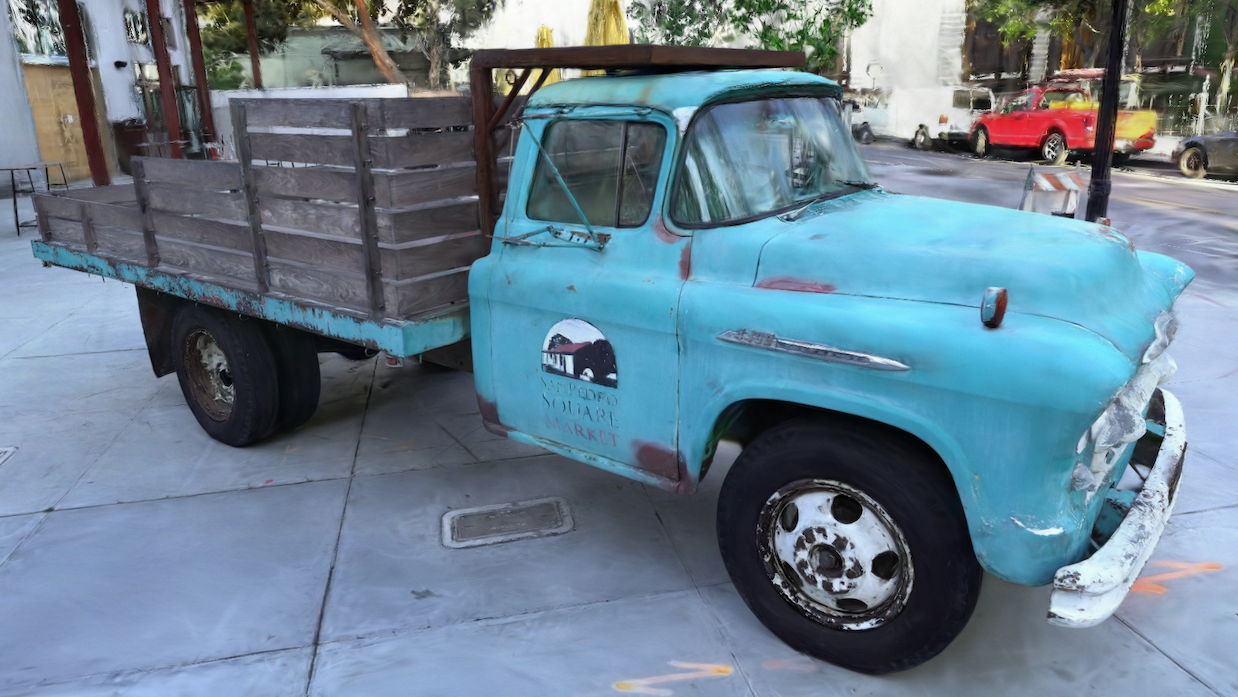} 
    \\
    \includegraphics[width=\imheight,height=\imheight]{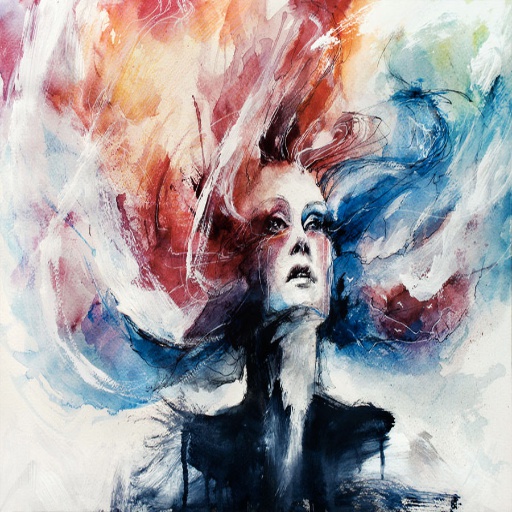} & 
    \includegraphics[width=\imwidth,height=\imheight]{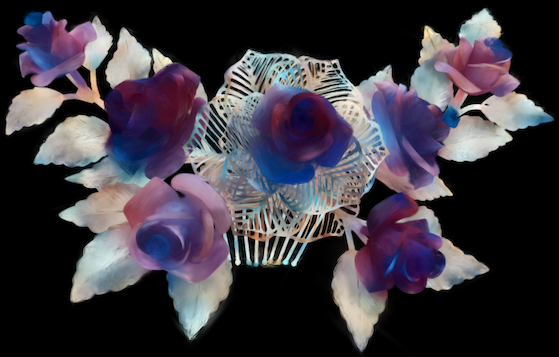} &
    \includegraphics[width=\imwidth,height=\imheight]{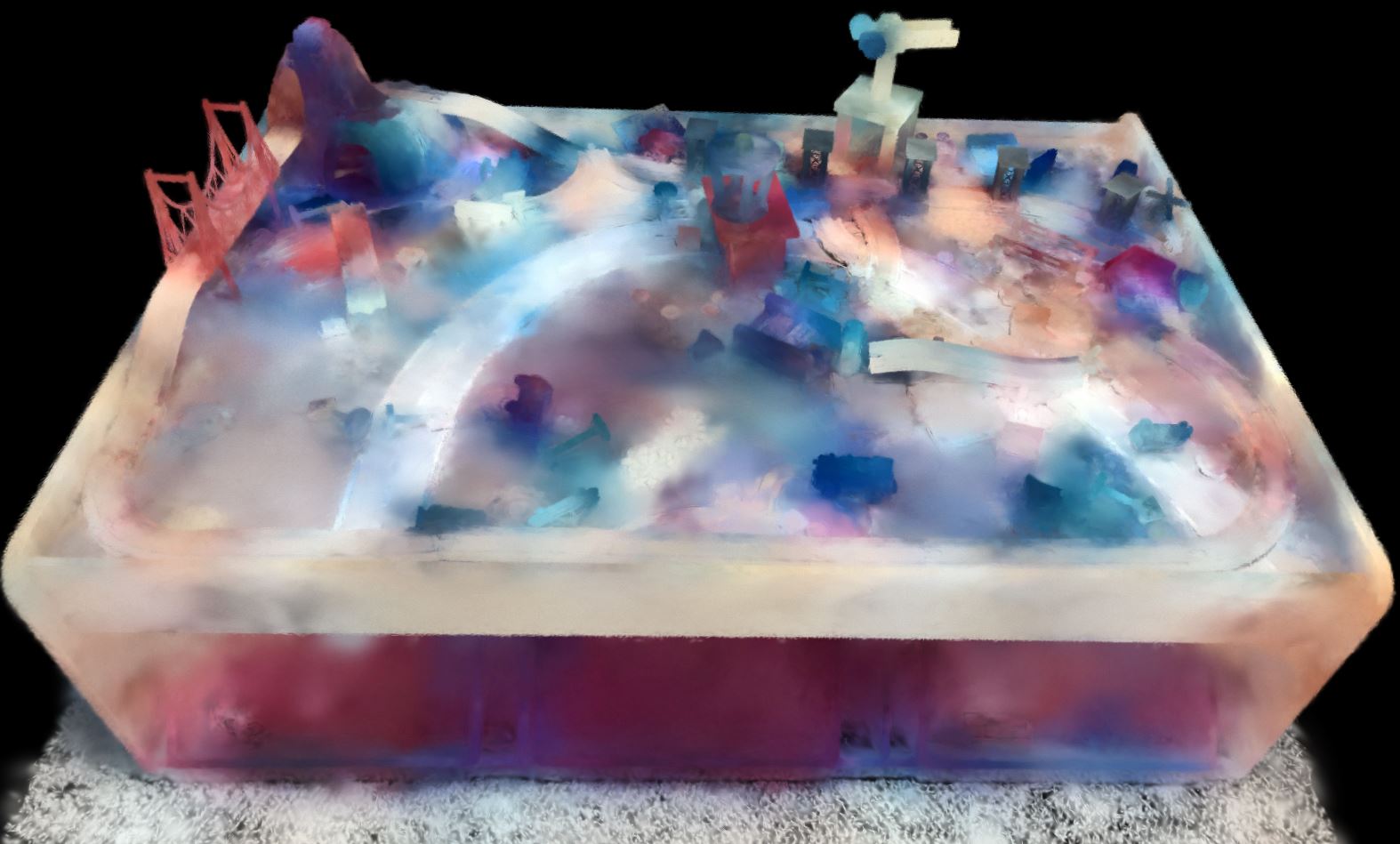} & 
    \includegraphics[width=\imwidth,height=\imheight]{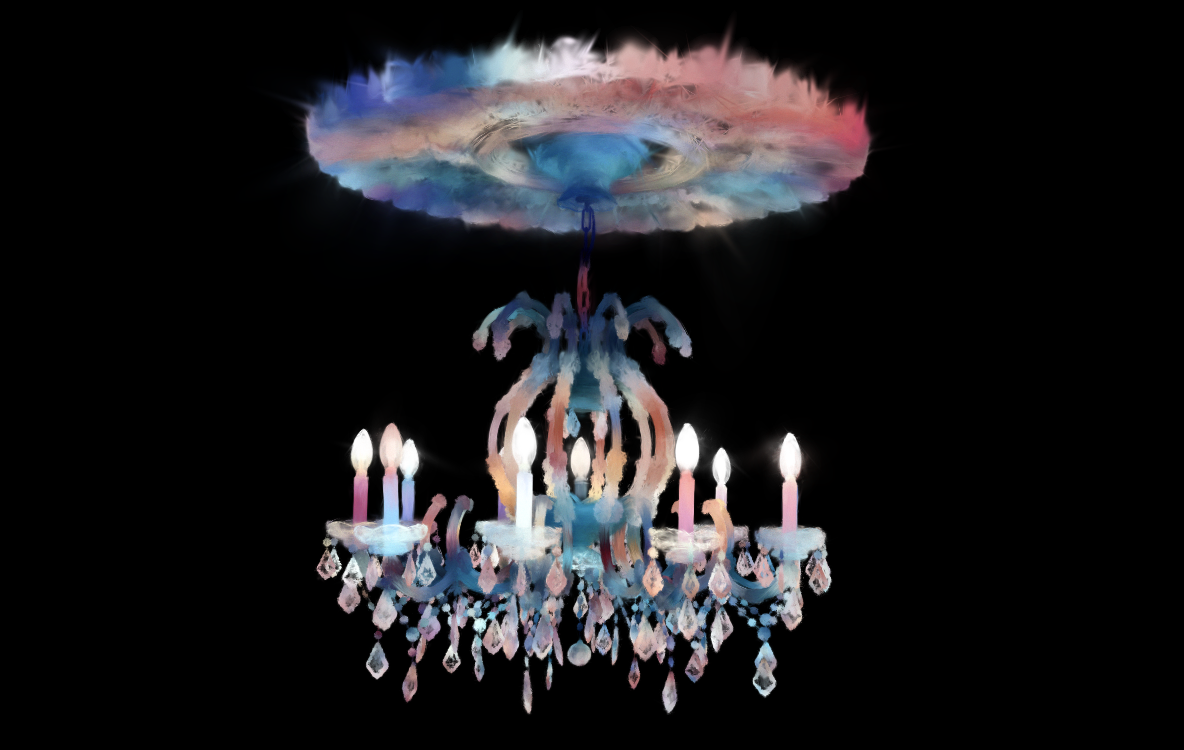} &
    \includegraphics[width=\imwidth,height=\imheight]{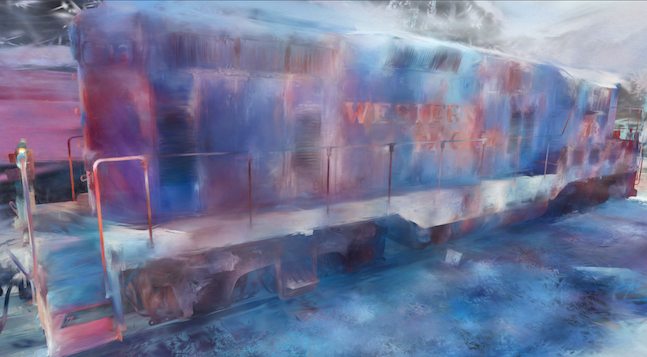} &  \includegraphics[width=\imwidth,height=\imheight]{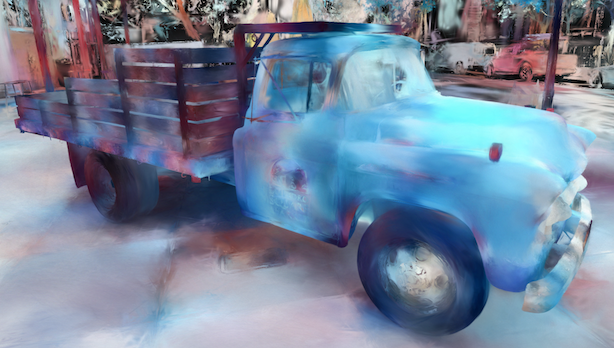}  
    \\
    \includegraphics[width=\imheight,height=\imheight]{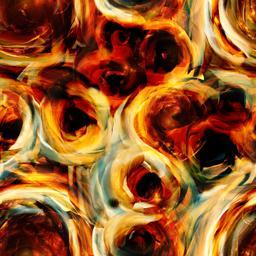}  & 
    \includegraphics[width=\imwidth,height=\imheight]{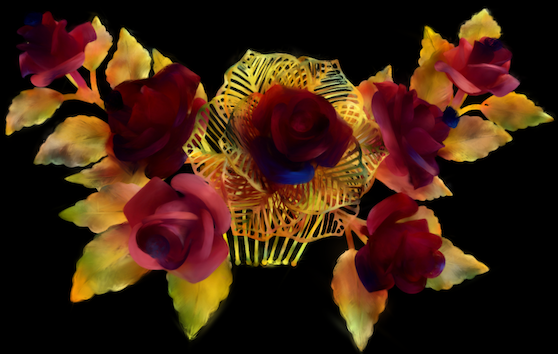} &
    \includegraphics[width=\imwidth,height=\imheight]{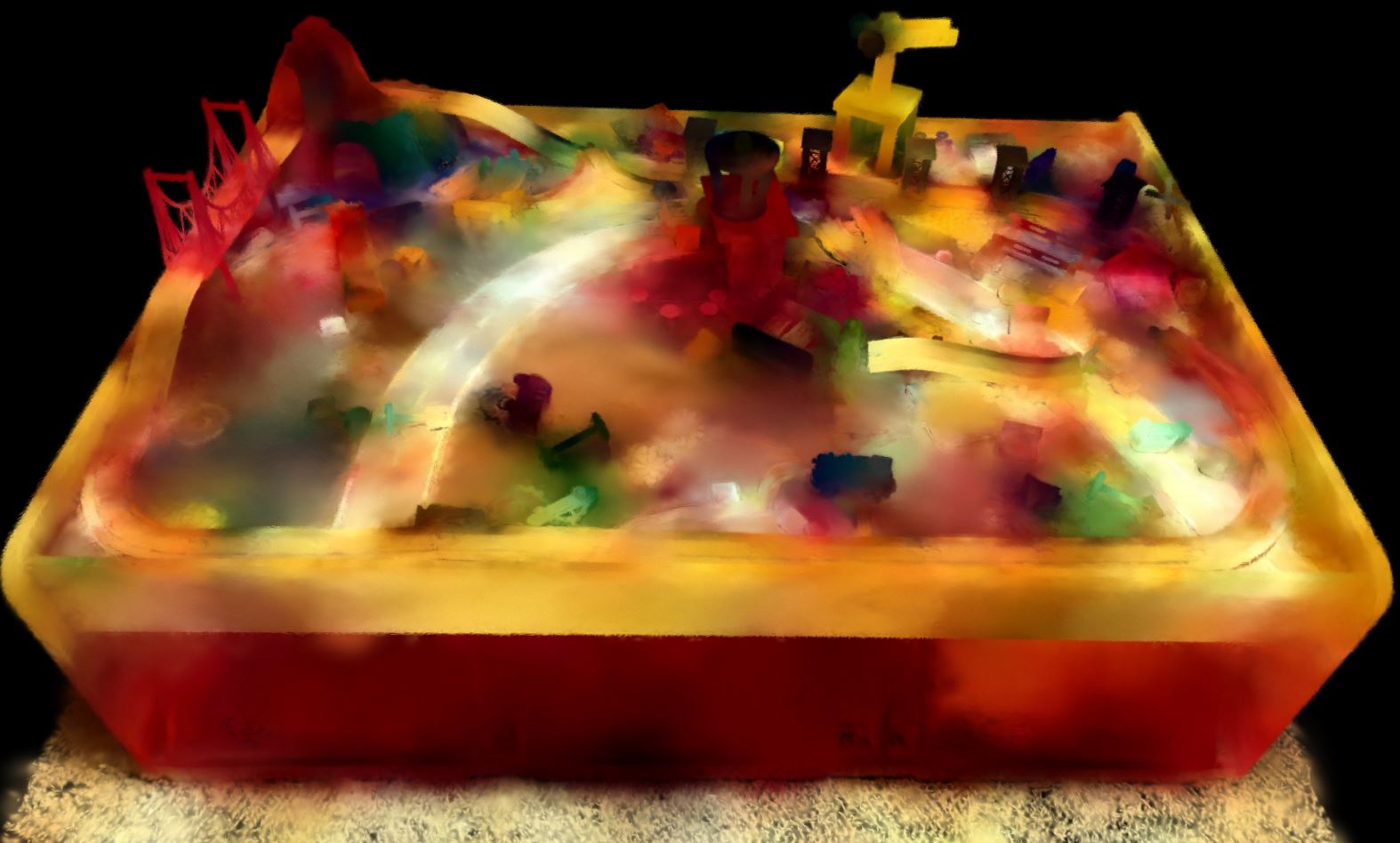} & 
    \includegraphics[width=\imwidth,height=\imheight]{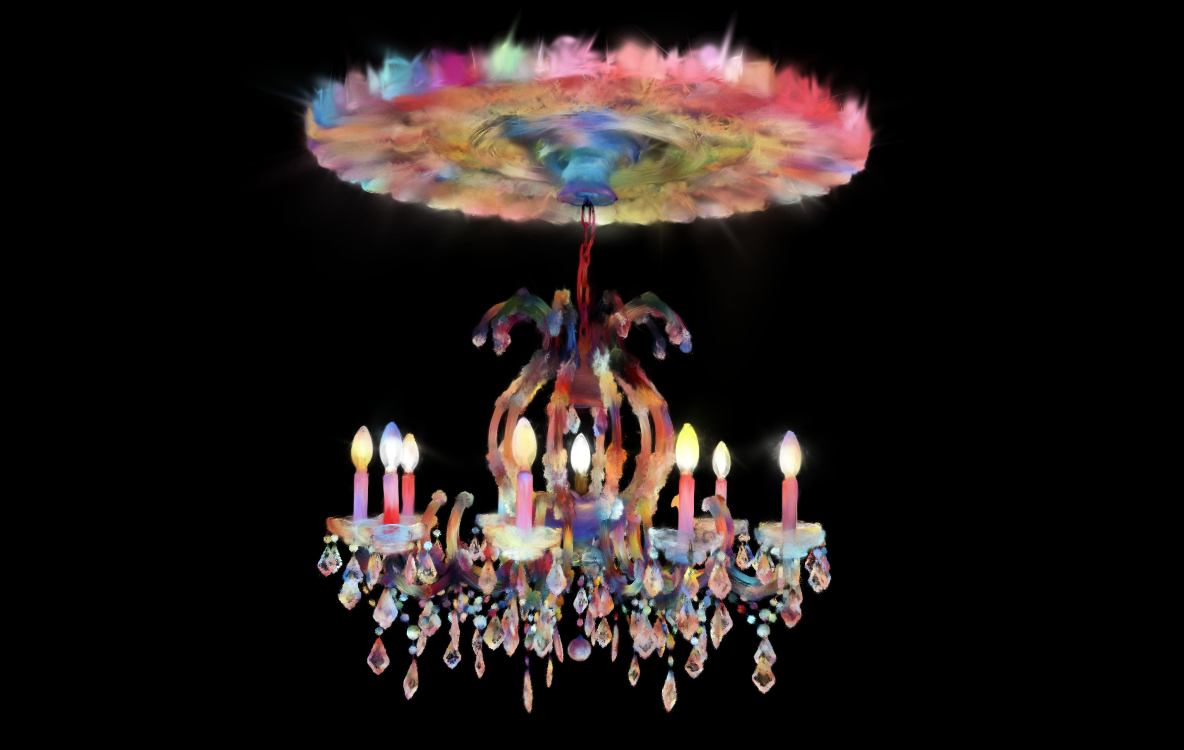} &
    \includegraphics[width=\imwidth,height=\imheight]{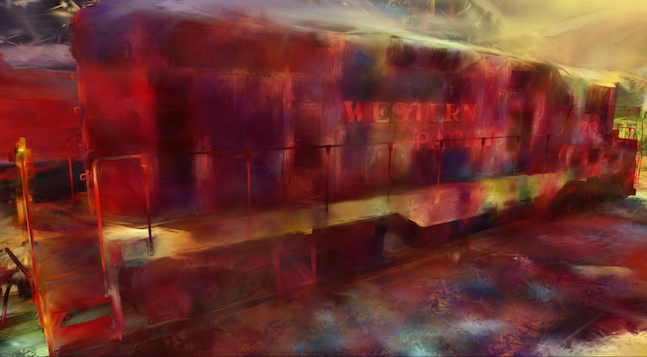} &  \includegraphics[width=\imwidth,height=\imheight]{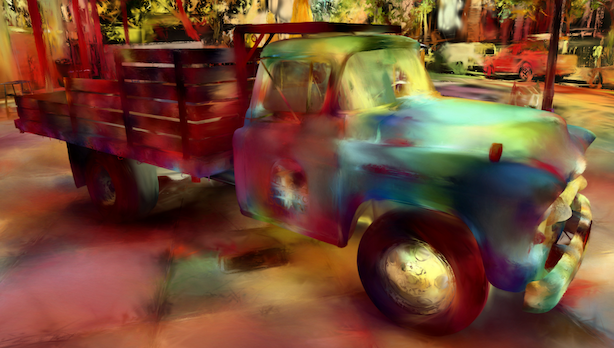} 
    \\
    \includegraphics[width=\imheight,height=\imheight]{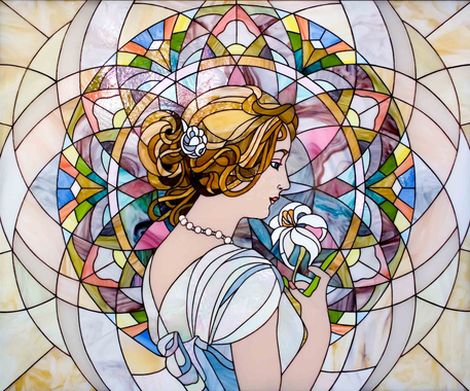}  & 
    \includegraphics[width=\imwidth,height=\imheight]{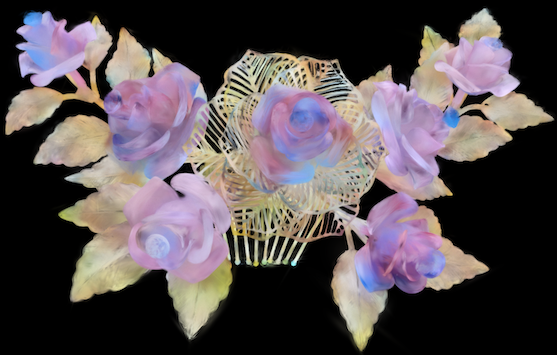} &
    \includegraphics[width=\imwidth,height=\imheight]{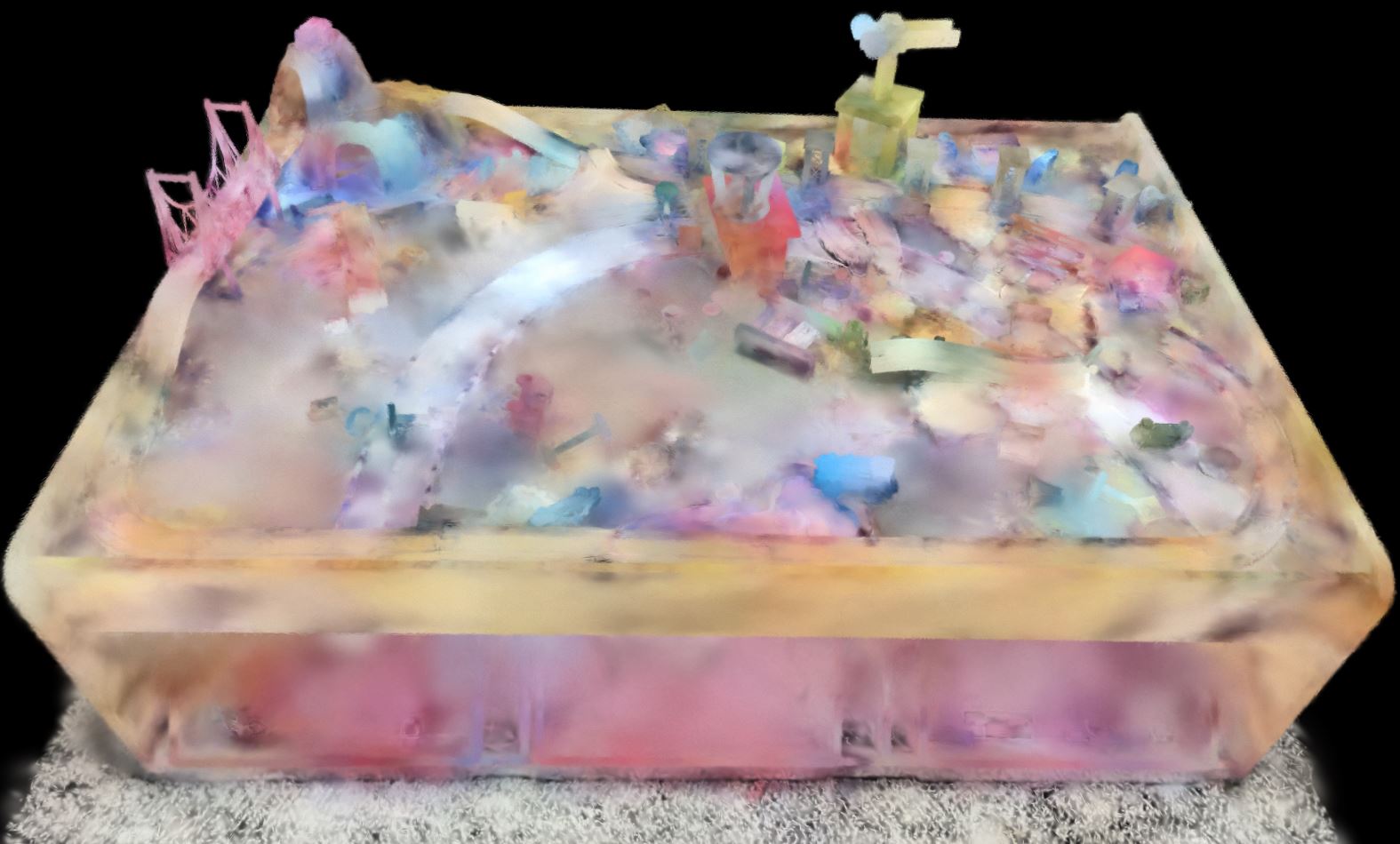}  & 
    \includegraphics[width=\imwidth,height=\imheight]{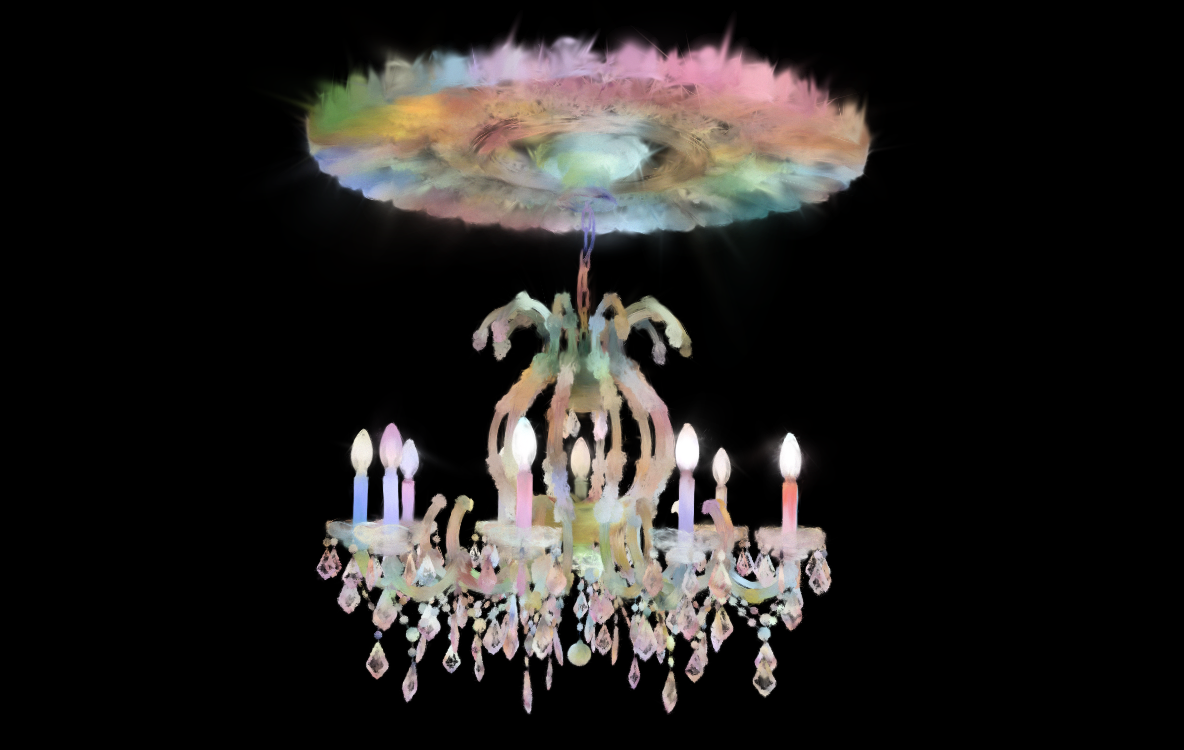} &
    \includegraphics[width=\imwidth,height=\imheight]{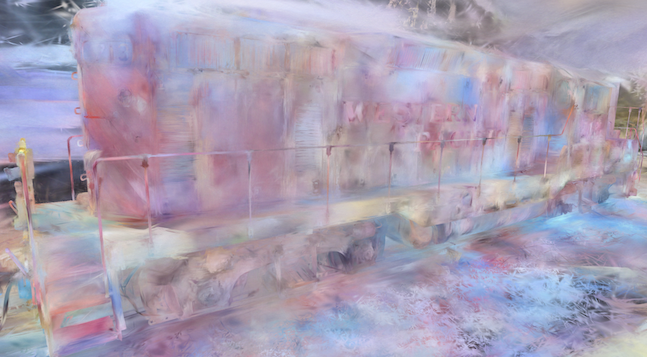} &  \includegraphics[width=\imwidth,height=\imheight]{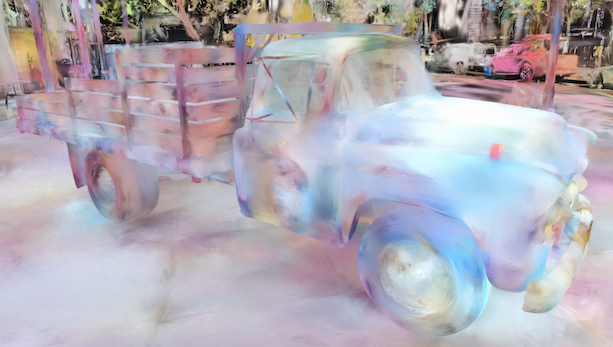} 
    \\
    \includegraphics[width=\imheight,height=\imheight]{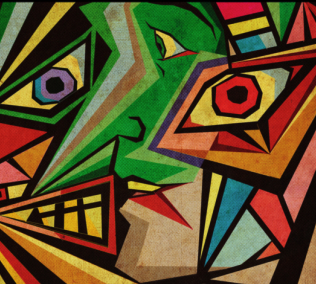}  & 
    \includegraphics[width=\imwidth,height=\imheight]{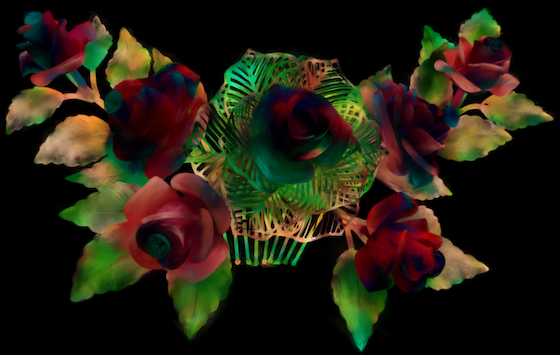} &
    \includegraphics[width=\imwidth,height=\imheight]{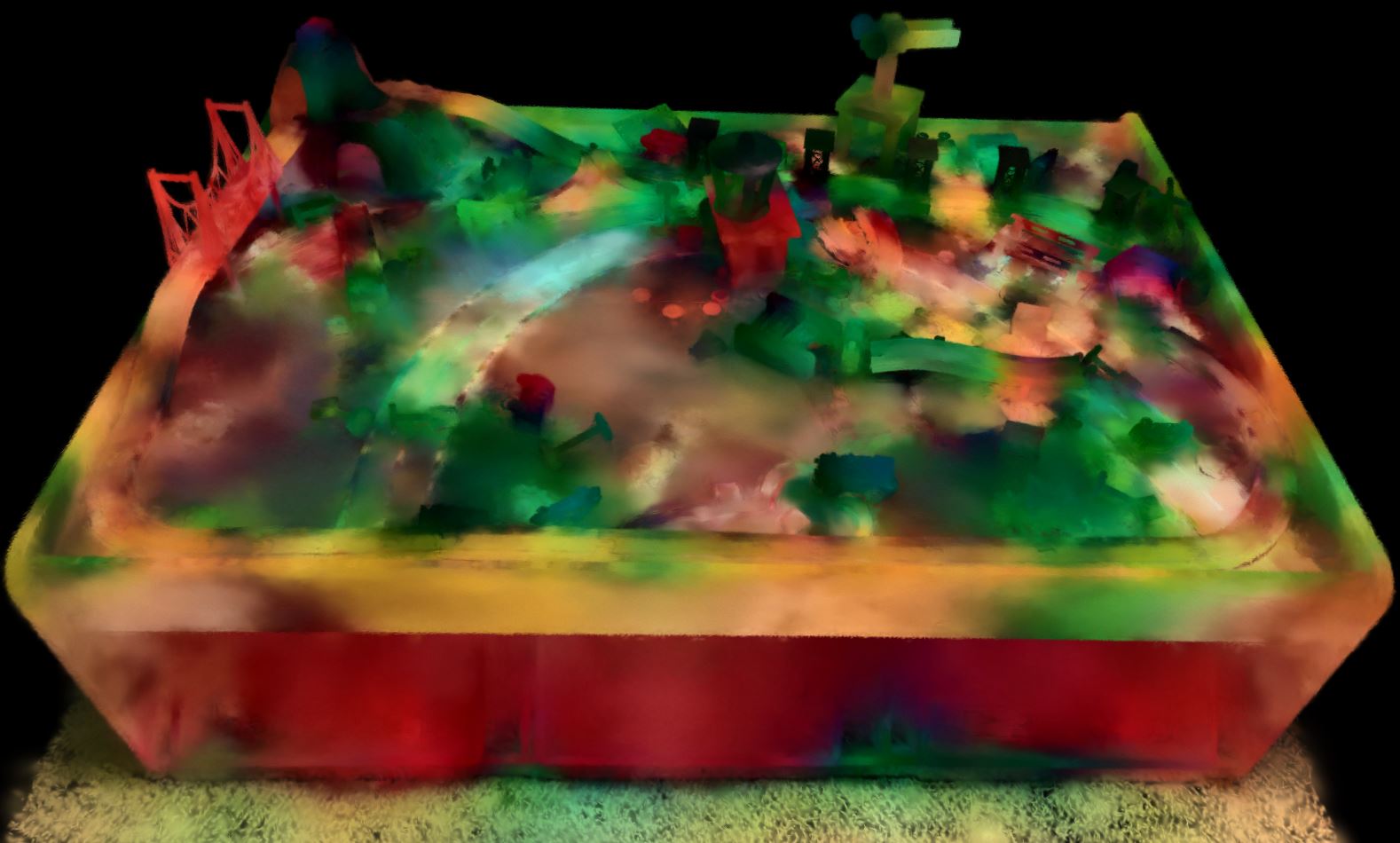} & 
    \includegraphics[width=\imwidth,height=\imheight]{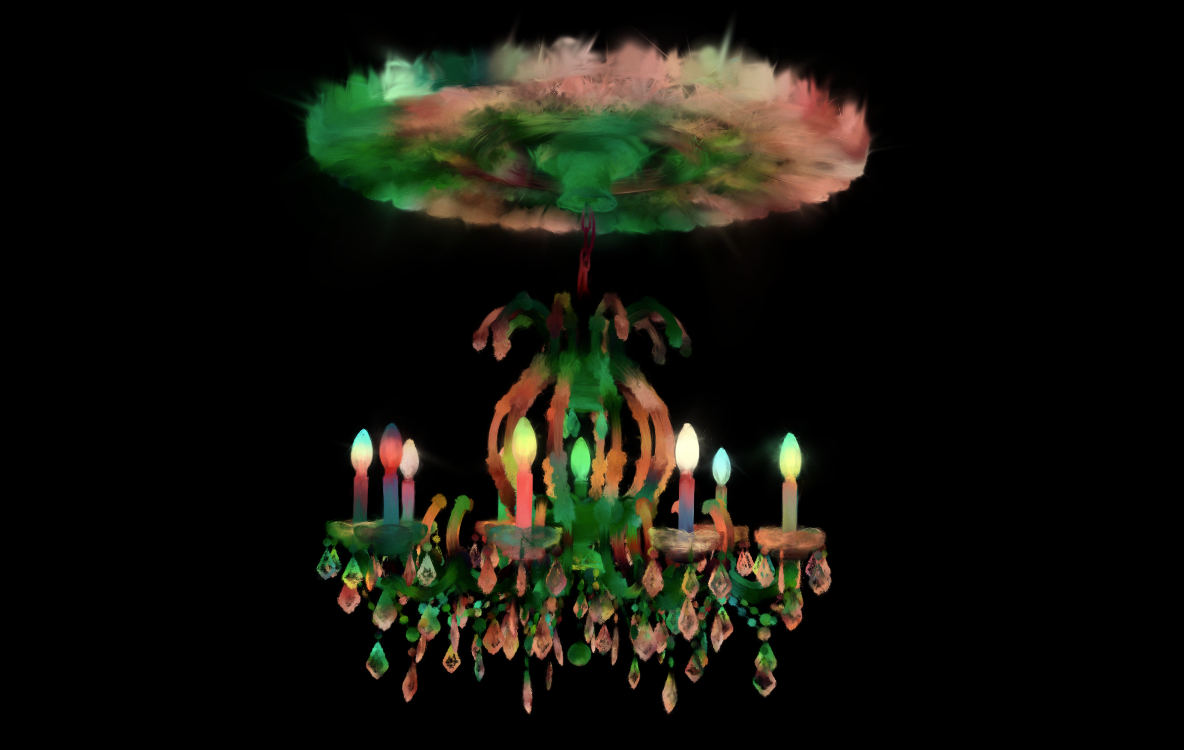} &
    \includegraphics[width=\imwidth,height=\imheight]{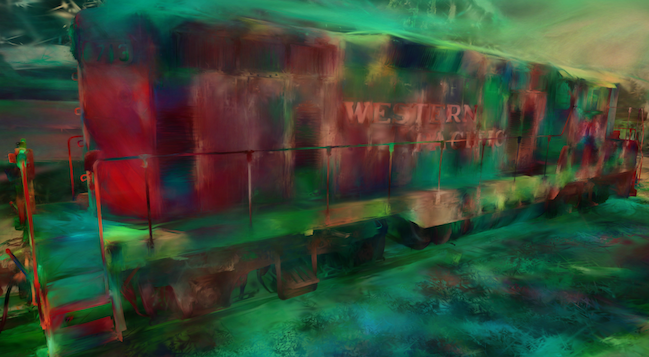} &  \includegraphics[width=\imwidth,height=\imheight]{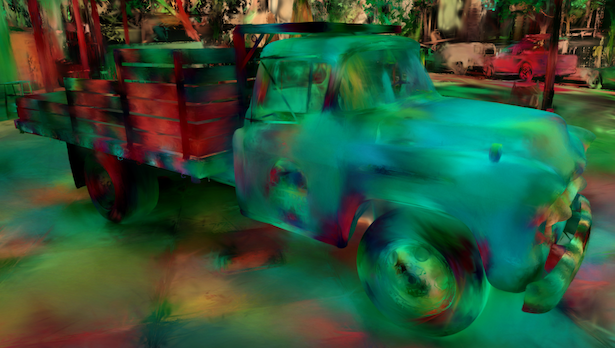} 
    \\
    \includegraphics[width=\imheight,height=\imheight]{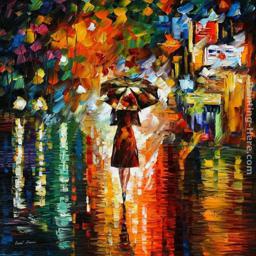}  & 
    \includegraphics[width=\imwidth,height=\imheight]{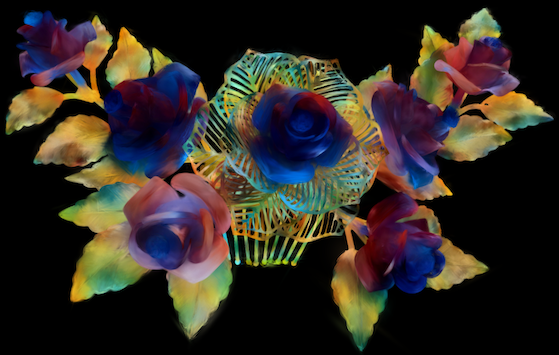} &
    \includegraphics[width=\imwidth,height=\imheight]{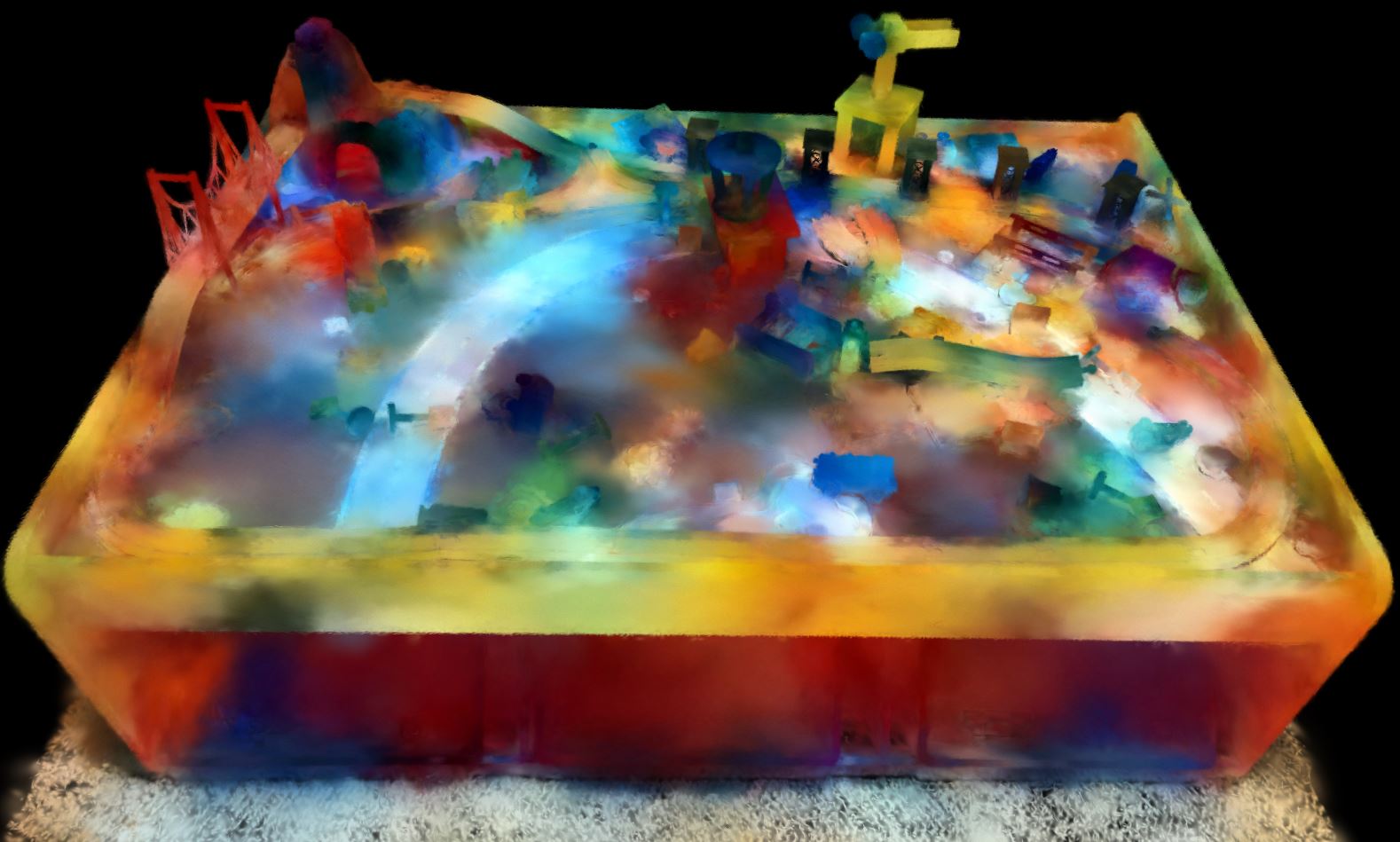} & 
    \includegraphics[width=\imwidth,height=\imheight]{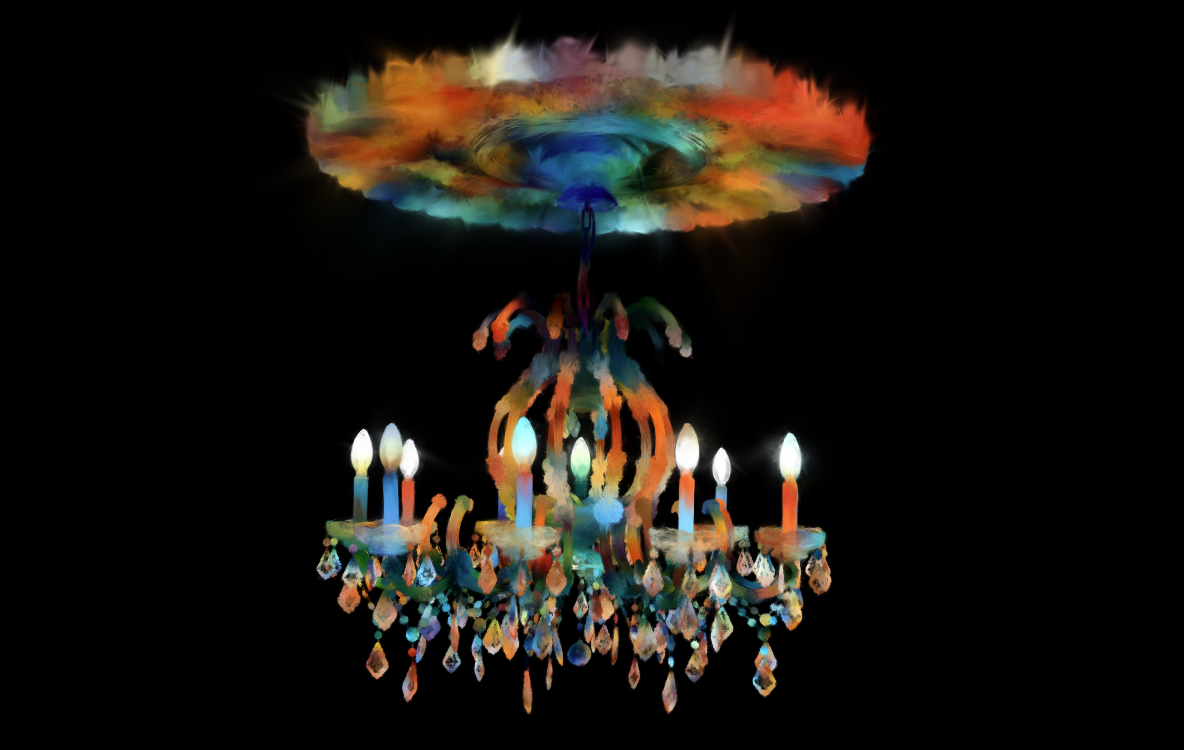} &
    \includegraphics[width=\imwidth,height=\imheight]{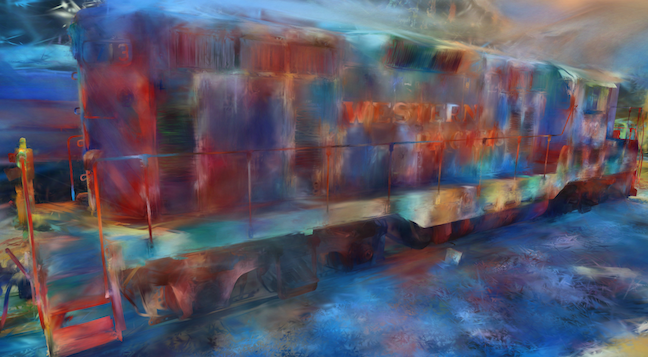} &  \includegraphics[width=\imwidth,height=\imheight]{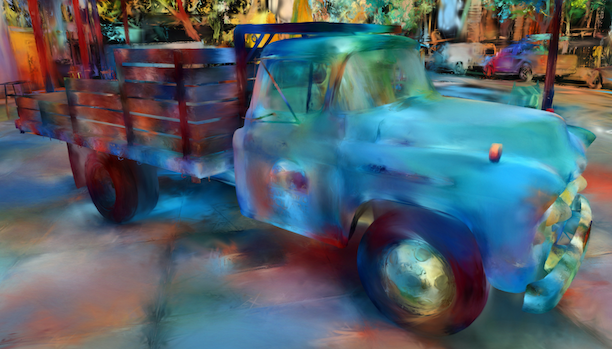} 
    \\
    \includegraphics[width=\imheight,height=\imheight]{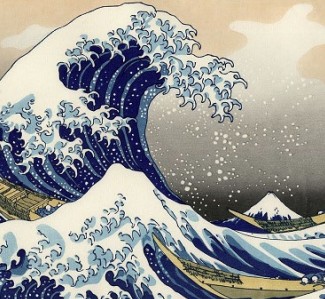}  & 
    \includegraphics[width=\imwidth,height=\imheight]{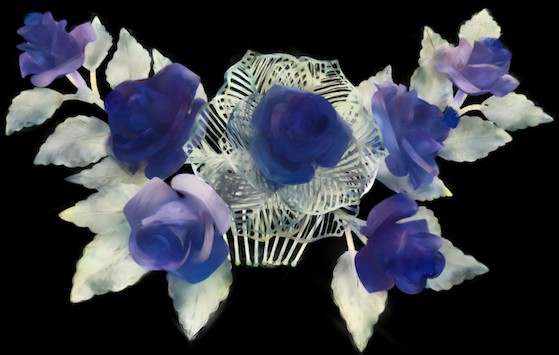}  &
    \includegraphics[width=\imwidth,height=\imheight]{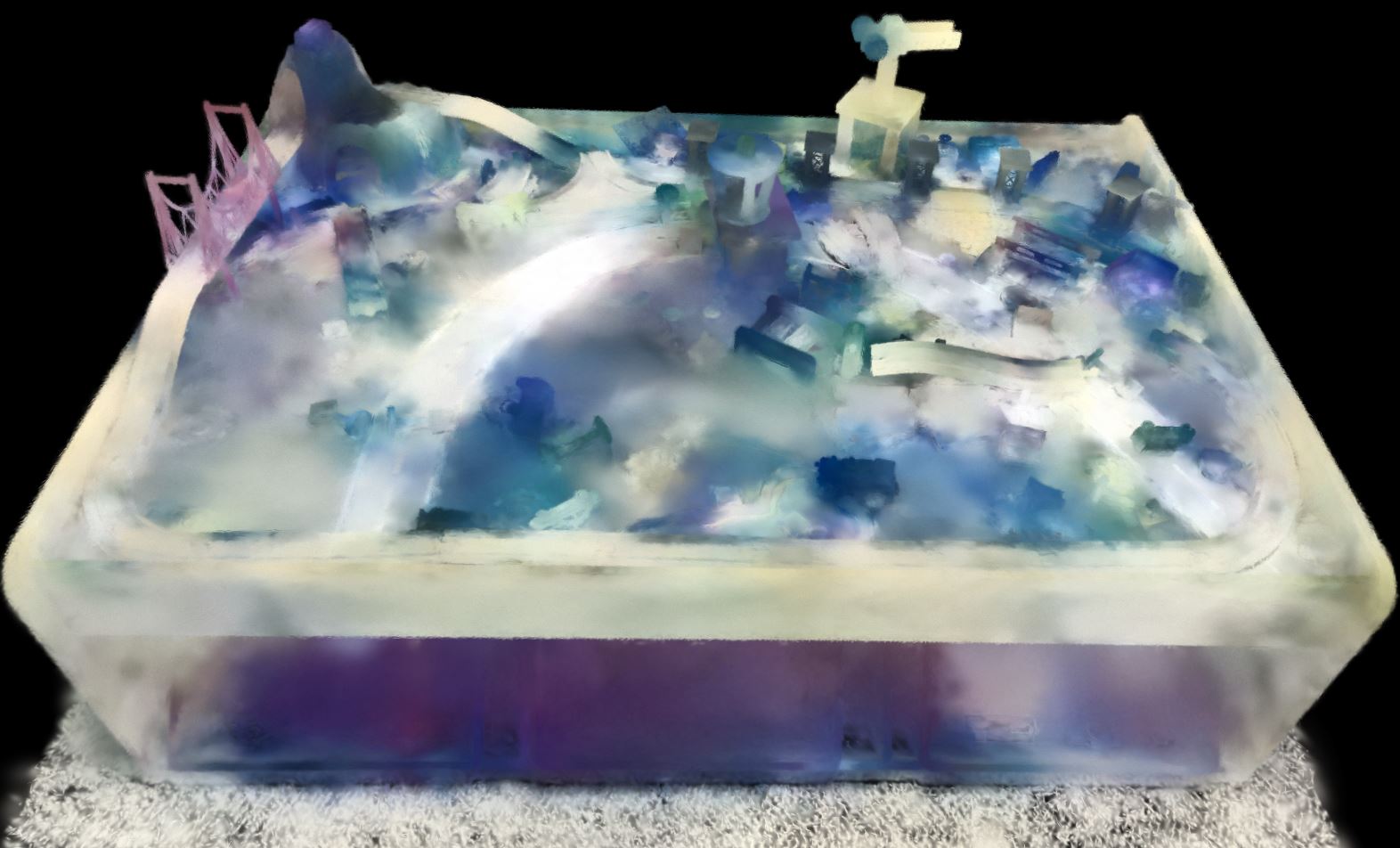}  & 
    \includegraphics[width=\imwidth,height=\imheight]{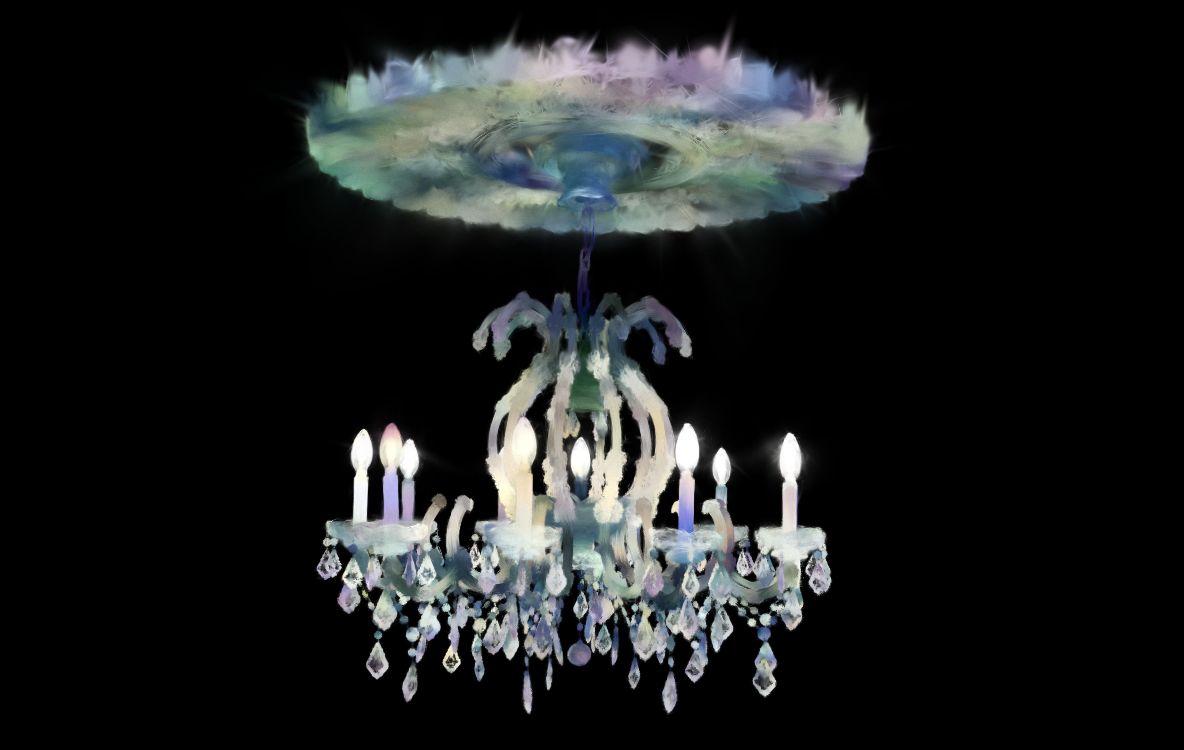} &
    \includegraphics[width=\imwidth,height=\imheight]{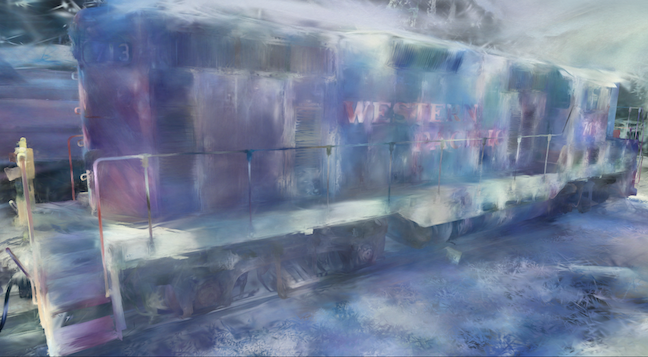} &  \includegraphics[width=\imwidth,height=\imheight]{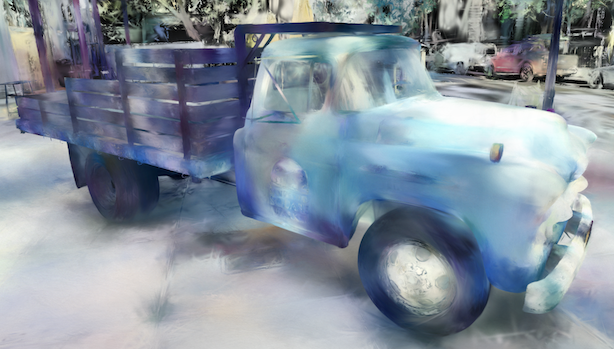} 
    \\
    Size: & 129669 & 286065 &  526504 & 1024454 & 2536143 
    \\
    After Sampling: &  175045 & 643489 & 809637 & 1733010 & 3084299 
    \\ 
    CPU Time: & 7.13s & 22.57s & 26.93s & 72.98s & 145.84s
    \\
    \end{tabular}
    \caption{Qualitative results of our approach. Example outputs are given for each of the style images along the left column and content 3DGS along the top row. The number of individual Gaussians (Size) and the number after adding additional samples is given along with its runtime on a Mac M2 CPU with no additional MPU or GPU acceleration.}
    \label{fig:style-grid}
\end{figure*}

The output stylizations of our approach for multiple styles and 3DGS scenes are given in Fig.~\ref{fig:style-grid}. Example 3DGS scenes were taken from the Tanks and Temples dataset \cite{Knapitsch2017}, public repositories \cite{kobranov2024splats}, and scans generated using the Scaniverse app \cite{Scaniverse}. 
High quality 3D Gaussian splats demonstrated excellent style transfer results, with good color alignment and excellent content preservation. Lower quality 3D Gaussian splats suffered at times from poor mapping of style elements to content and regions of sparse data.

Another advantage of our stylization method is its short runtime on consumer-grade hardware. 
For each of the examples shown in Fig.~\ref{fig:style-grid}, we indicate the time to complete the full stylization. These times are from a consumer-grade Mac M2 CPU with no MPU or GPU acceleration. The examples are sorted from smallest to largest, illustrating how the runtime increase as 3DGS size increases. As is shown, our method runs in sub minute times on smaller splat sizes, and runs in approximately 2 minutes on large splat scenes. Again, no camera parameters, COLMAP outputs, or rendered views are needed for the stylization, allowing for direct use of \textit{.splat} or \textit{.ply} files as inputs into the model. It also saves files directly back to \textit{.splat} files so that the stylization is end-to-end and viewer independent. This makes our approach immediately applicable to content creation pipelines and integration within 3DGS editing tools.

\newcommand{\imwidthc}{1.5in}
\newcommand{\imheightc}{0.90in}

\subsection{Comparisons to Other Methods}

\begin{figure*}[t!]
    \centering
    \setlength{\tabcolsep}{1.5pt}
    \begin{tabular}{ccccc}
         Style & Original & g-style \cite{kovacs2024} & StyleGaussian \cite{liu2024stylegaussian} & Ours\\
    
         \multirow{2}{*}{\includegraphics[width=0.65in,height=0.65in]{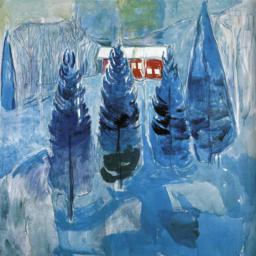}} 
         &
         \includegraphics[width=\imwidthc,height=\imheightc]{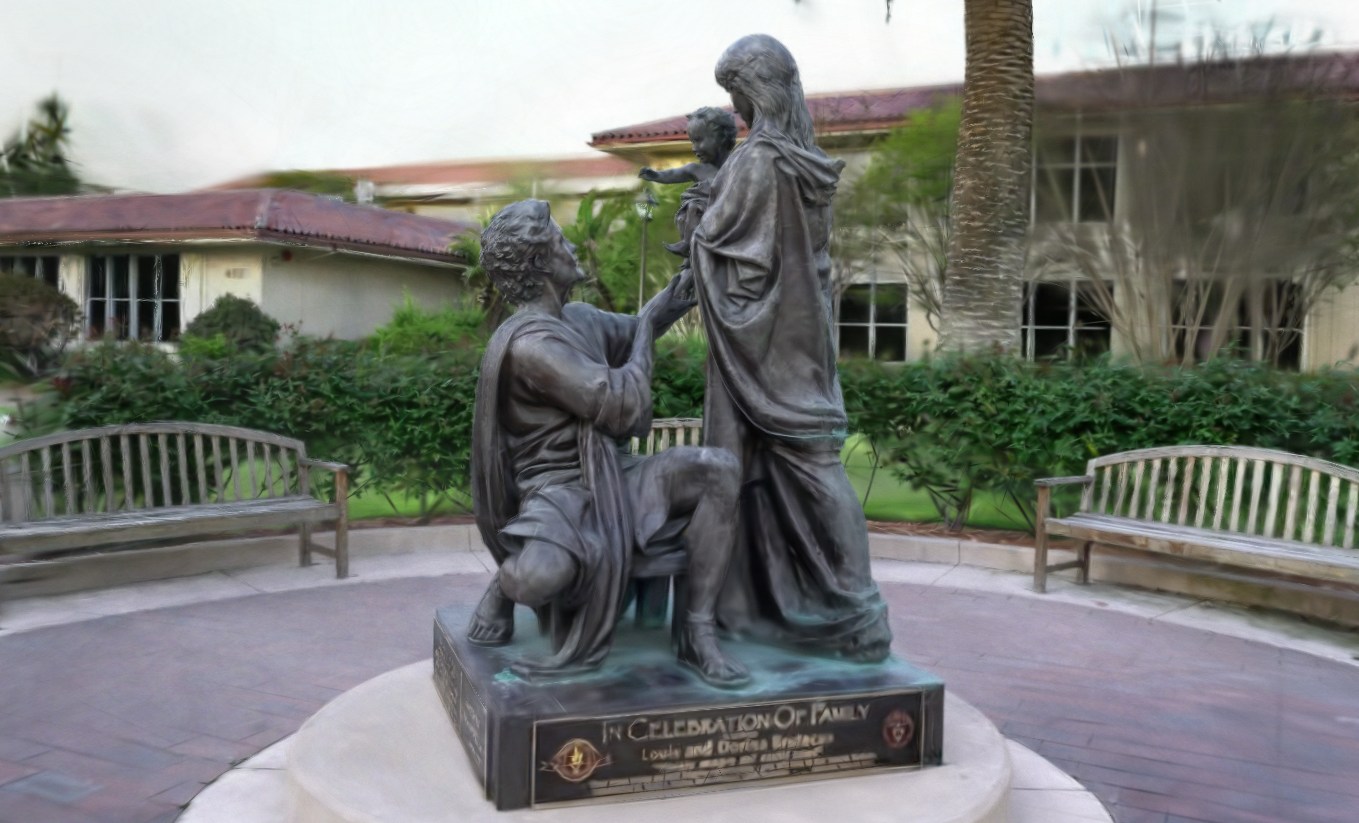} &
         \includegraphics[width=\imwidthc,height=\imheightc]{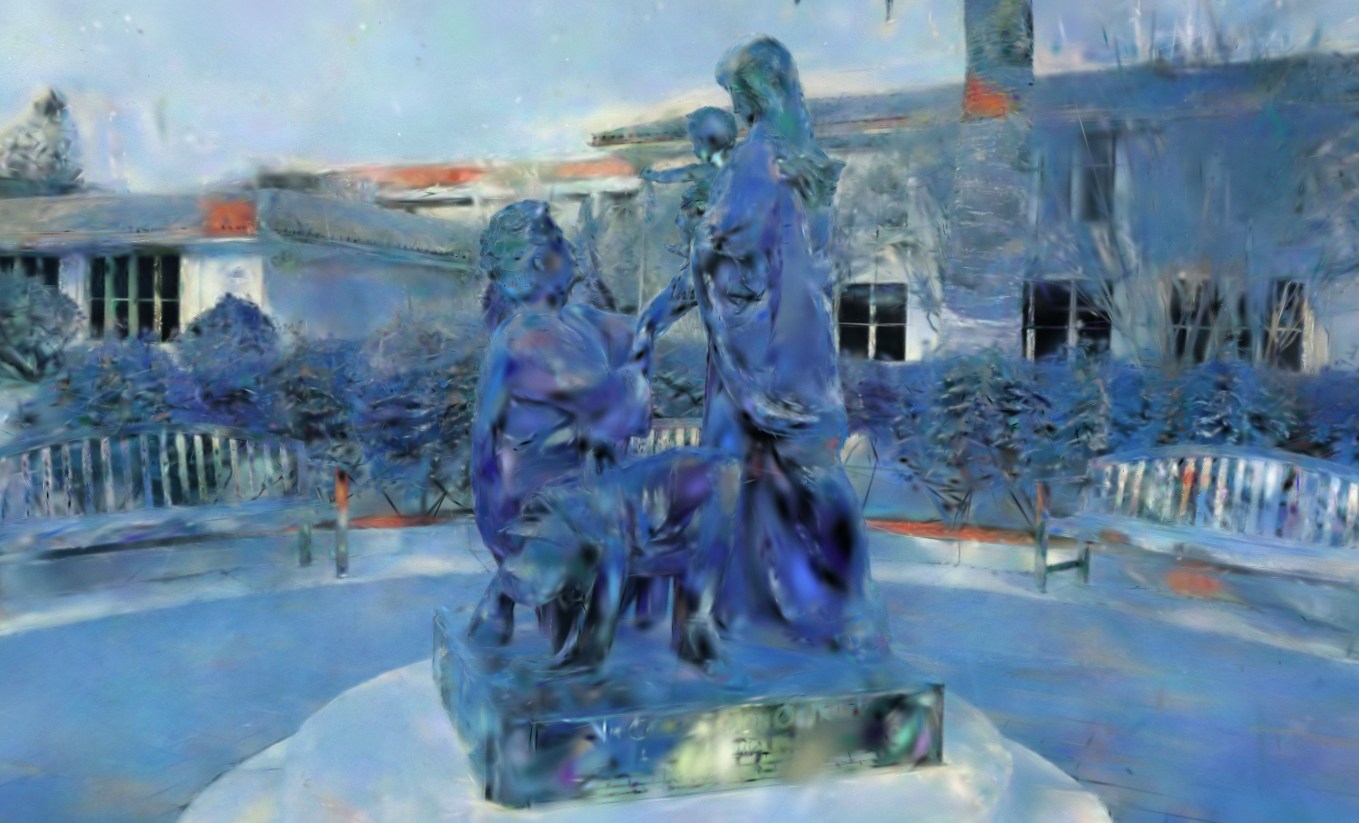} &
         \includegraphics[width=\imwidthc,height=\imheightc]{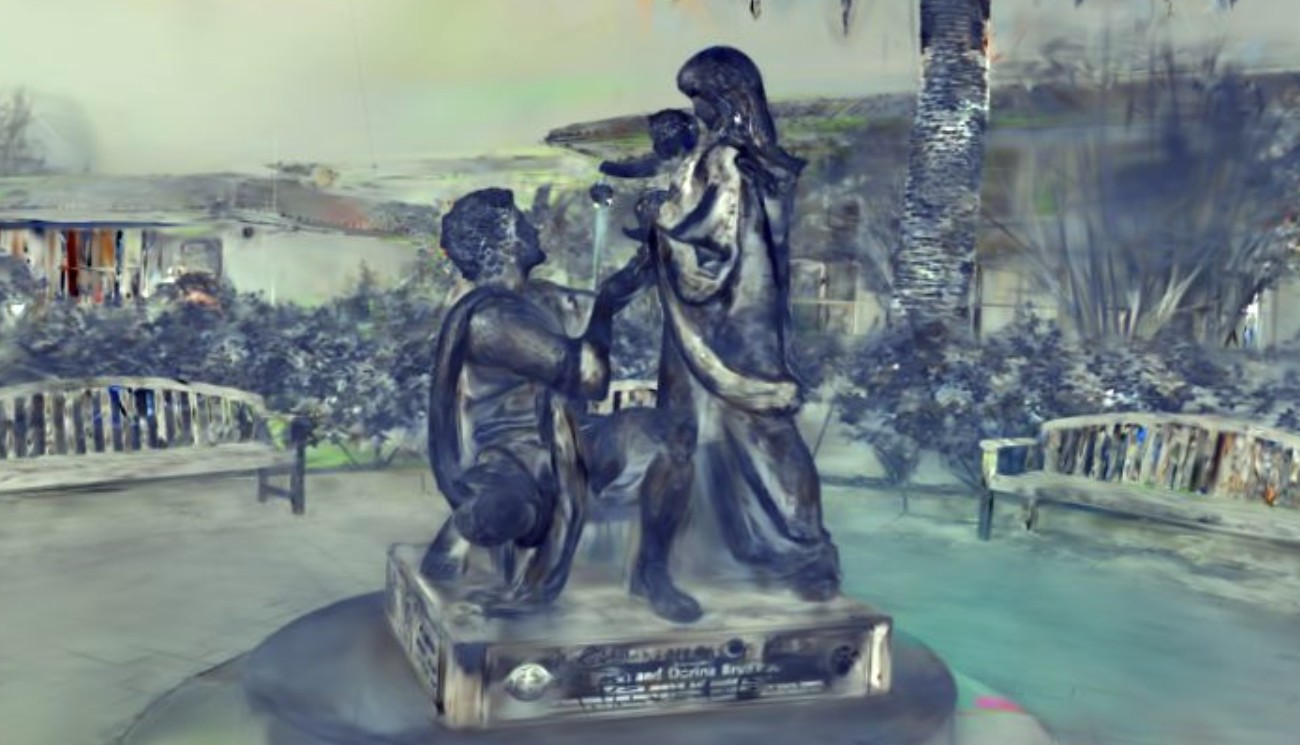} &
         \includegraphics[width=\imwidthc,height=\imheightc]{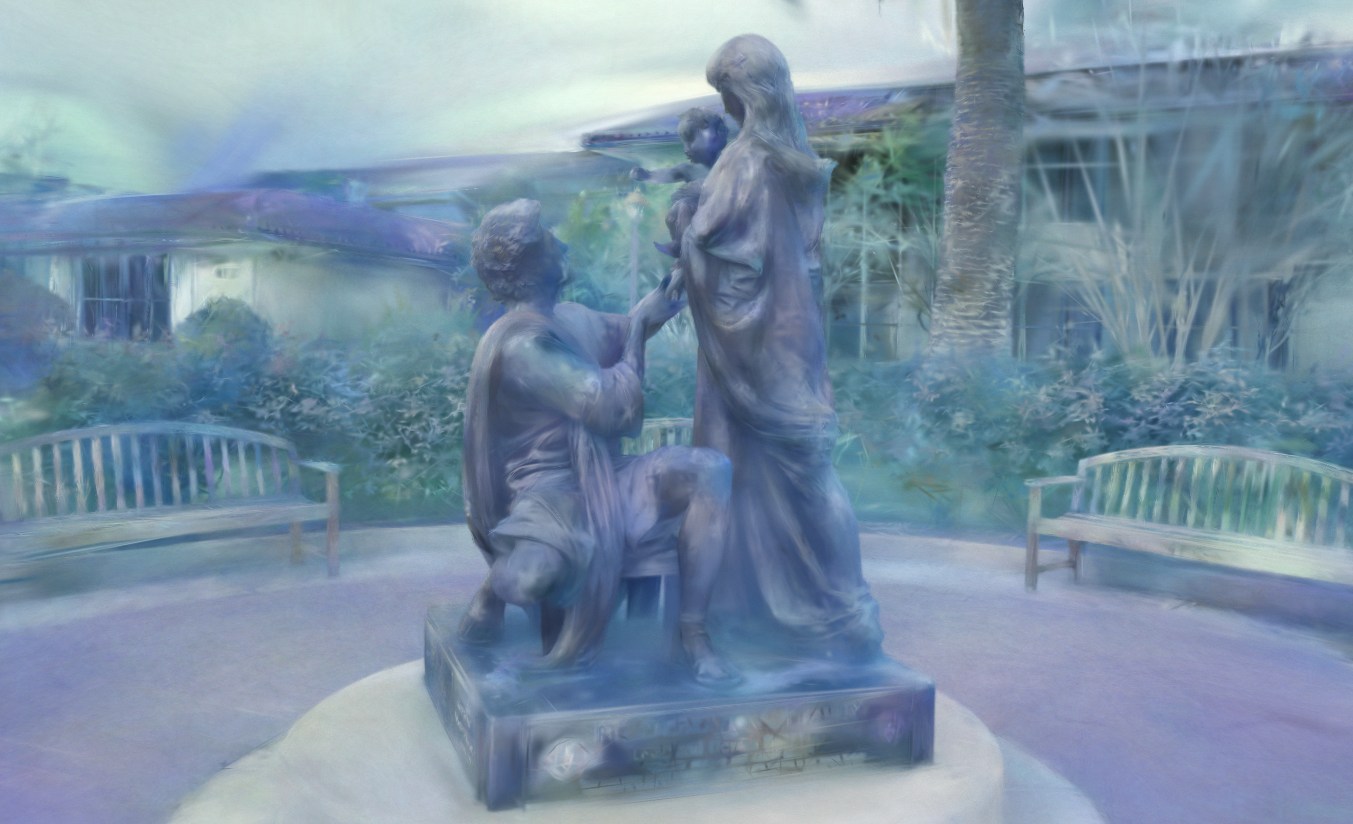} \\
         
          &
         \includegraphics[width=\imwidthc,height=\imheightc]{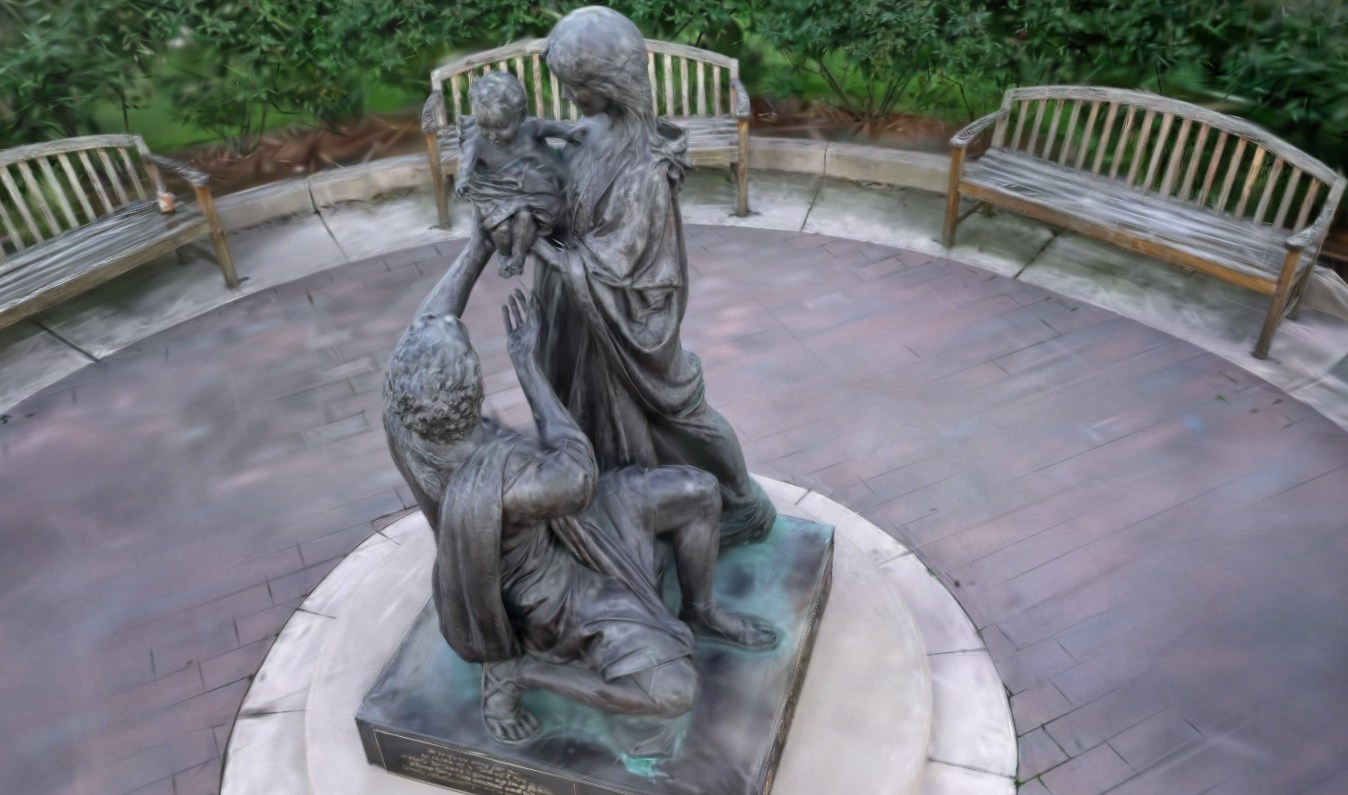} &
         \includegraphics[width=\imwidthc,height=\imheightc]{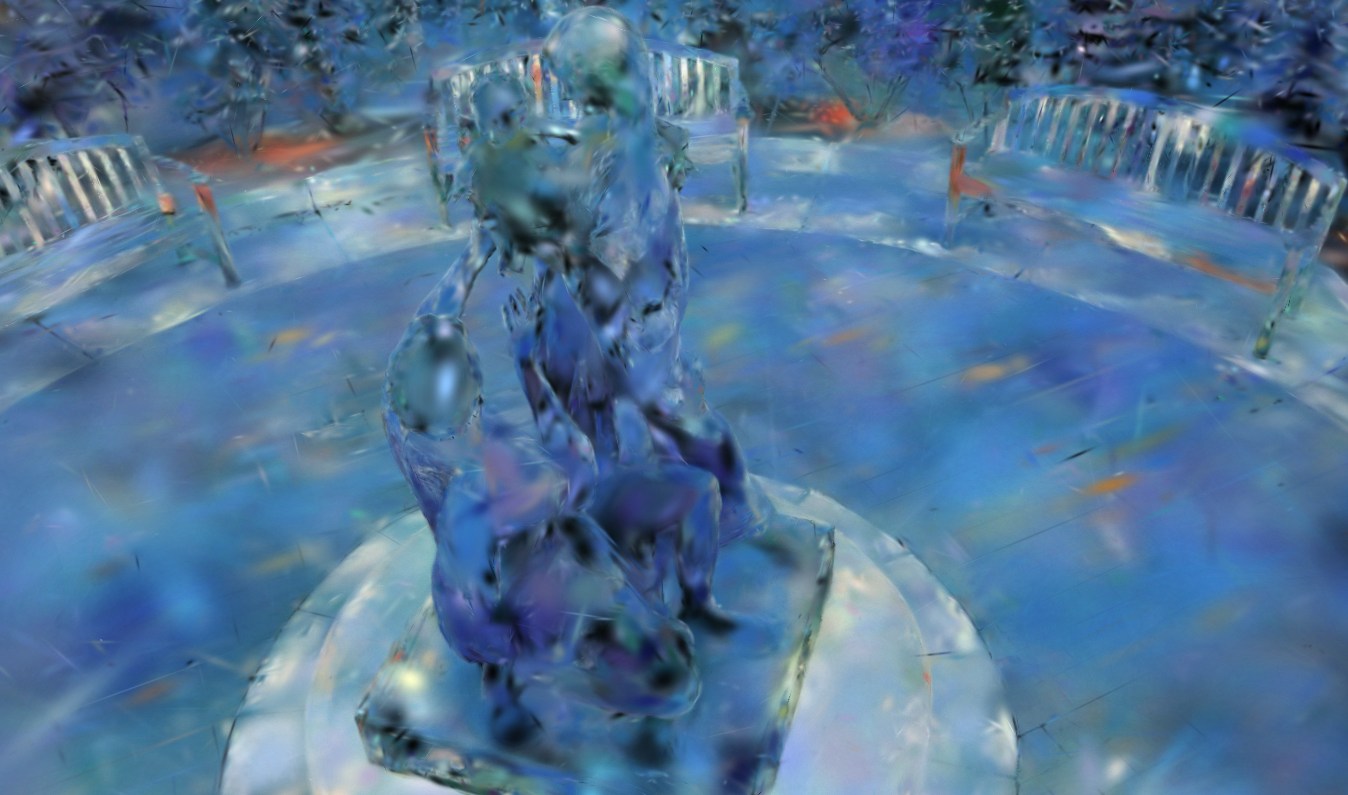} &
         \includegraphics[width=\imwidthc,height=\imheightc]{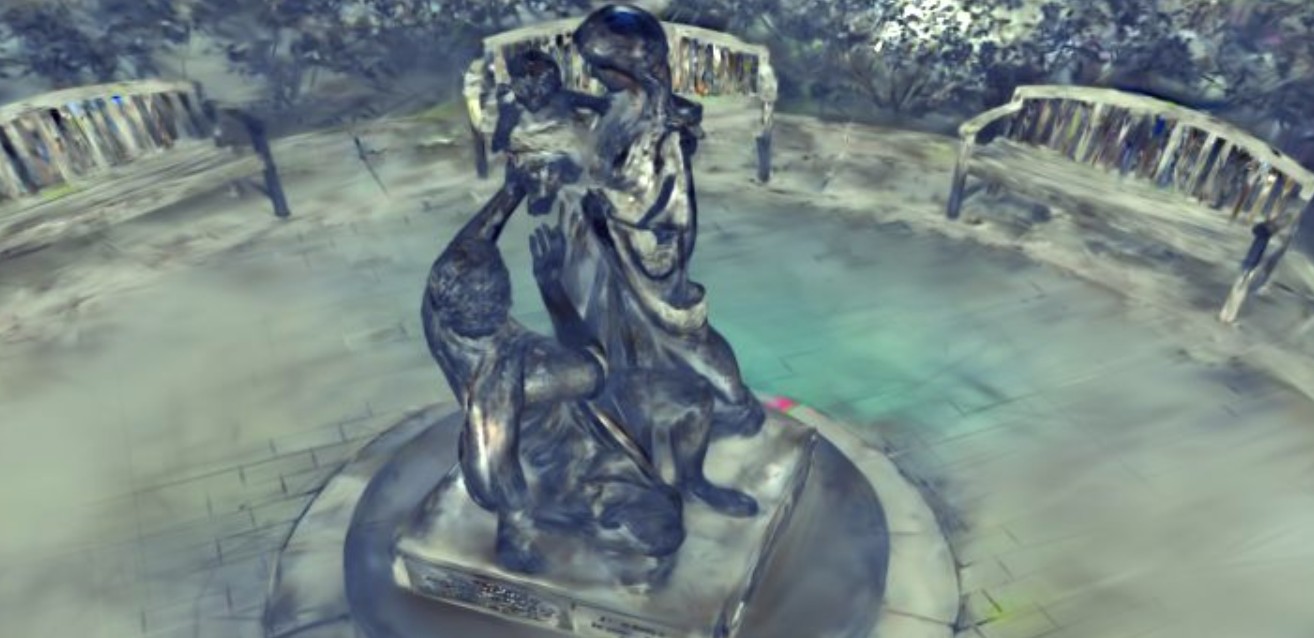} &
         \includegraphics[width=\imwidthc,height=\imheightc]{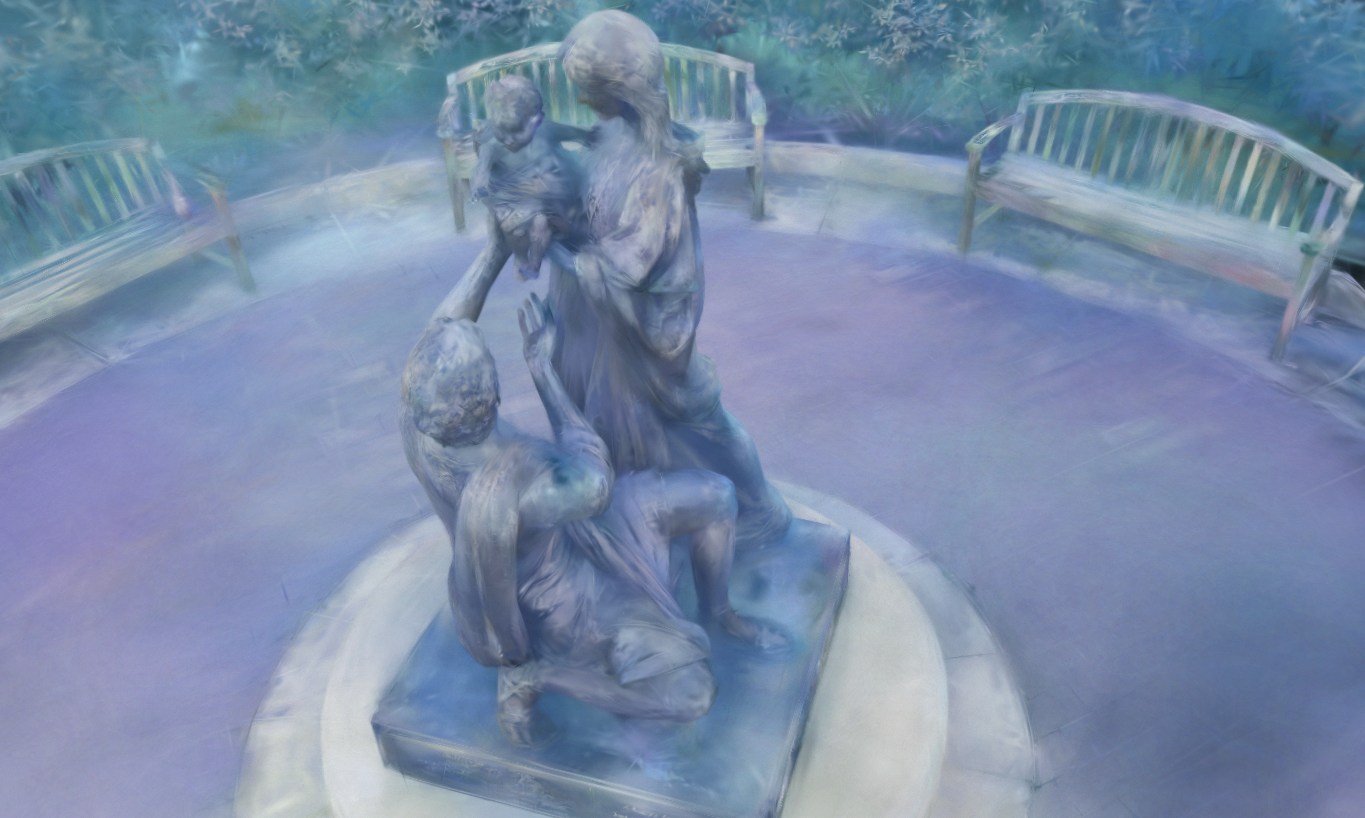} 
         \\

          & \multicolumn{1}{r}{LPIPS:} & 0.612 & 0.651 & 0.567
          \vspace{0.06in}
          \\
         \multirow{2}{*}{\includegraphics[width=0.65in,height=0.65in]{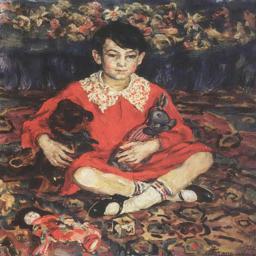}} 
         &
         \includegraphics[width=\imwidthc,height=\imheightc]{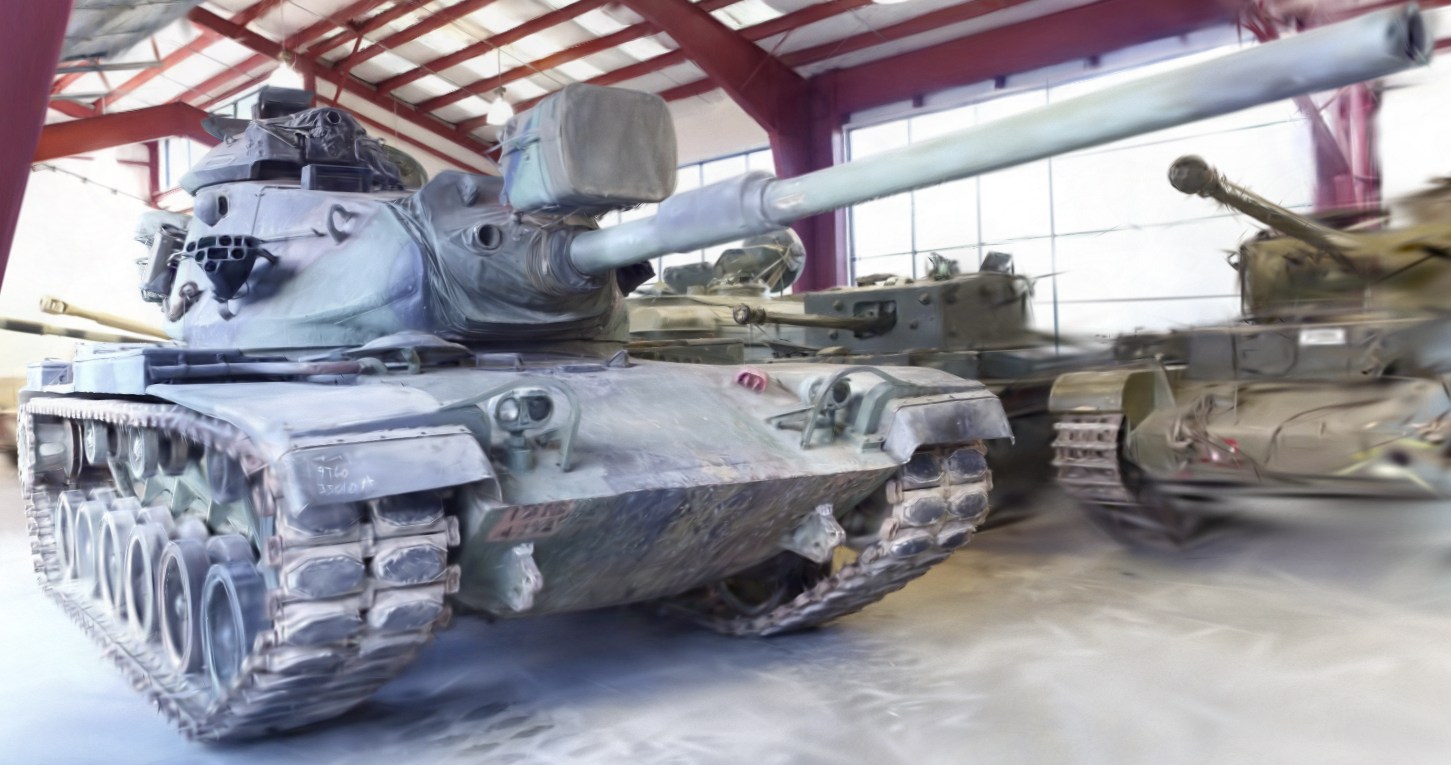} &
         \includegraphics[width=\imwidthc,height=\imheightc]{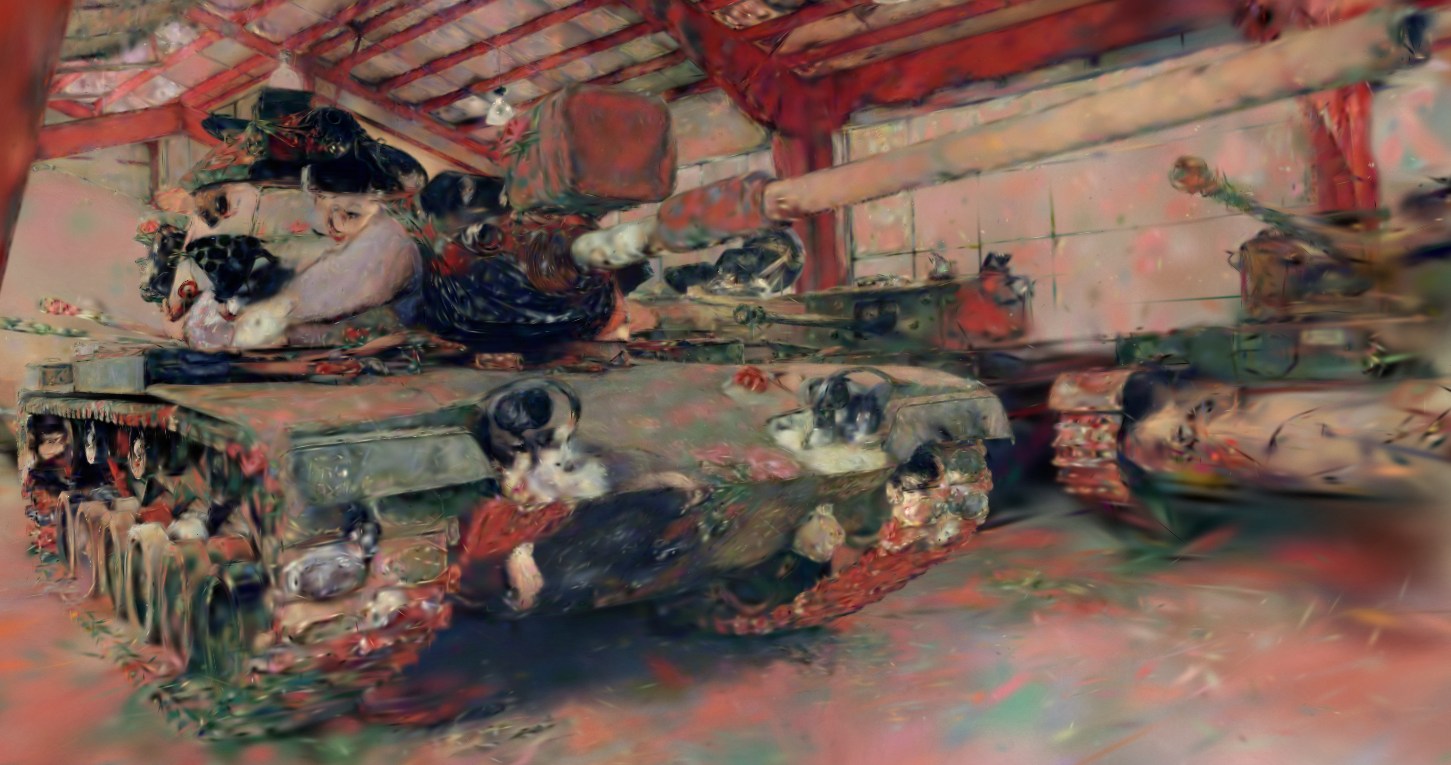} &
         \includegraphics[width=\imwidthc,height=\imheightc]{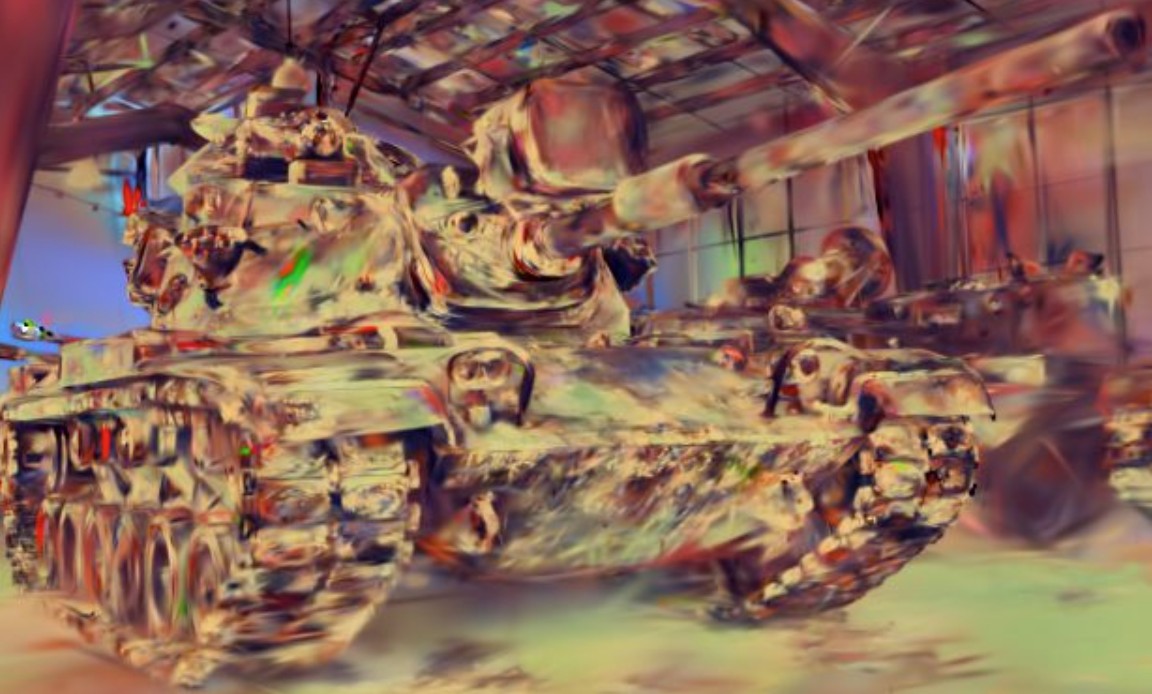} &
         \includegraphics[width=\imwidthc,height=\imheightc]{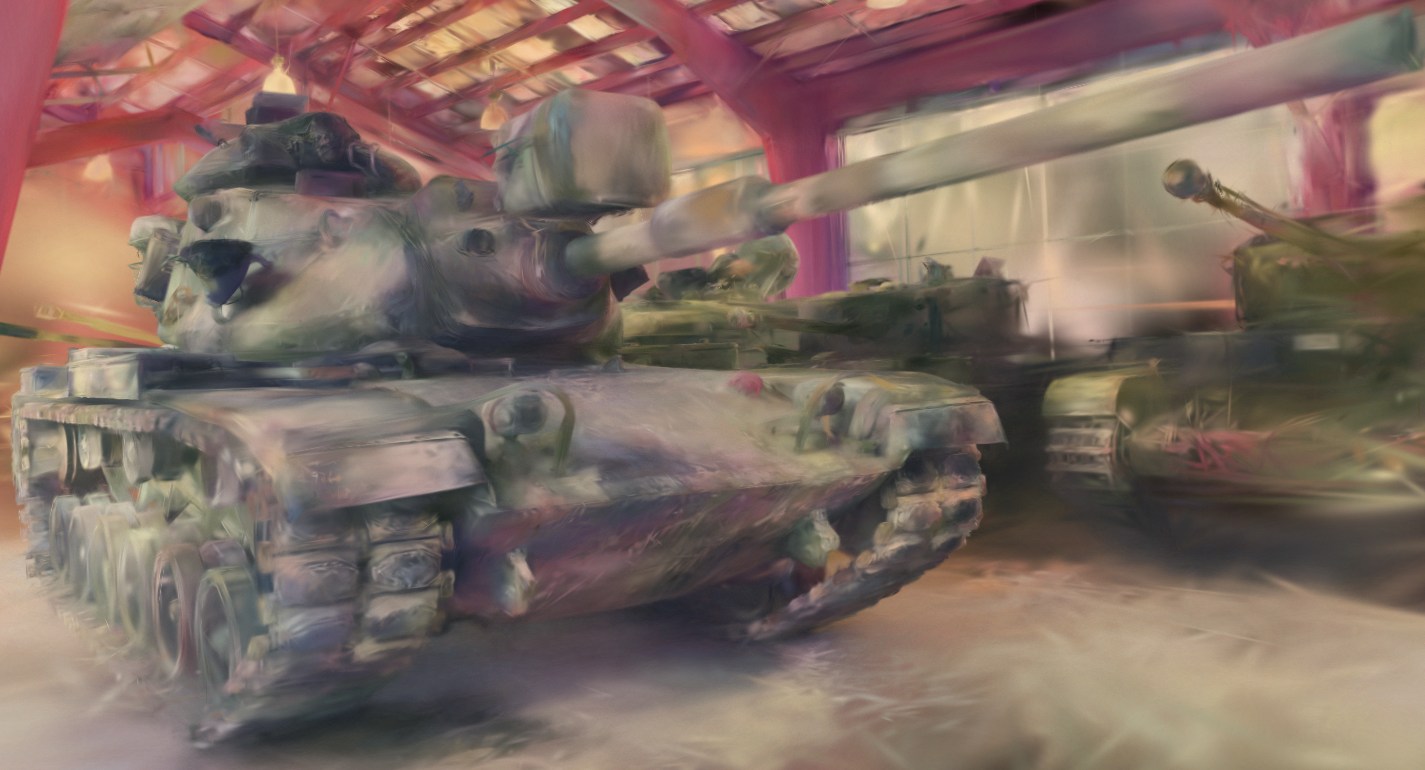} \\
         
         &
         \includegraphics[width=\imwidthc,height=\imheightc]{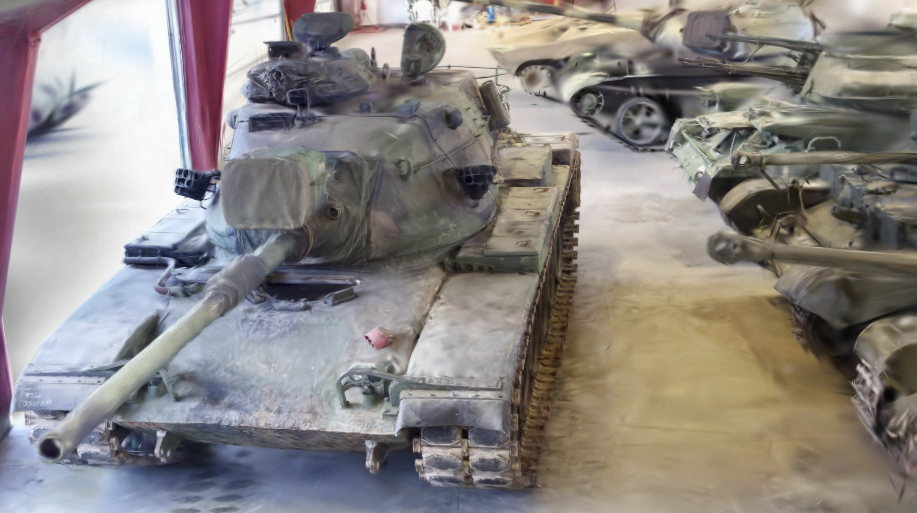} &
         \includegraphics[width=\imwidthc,height=\imheightc]{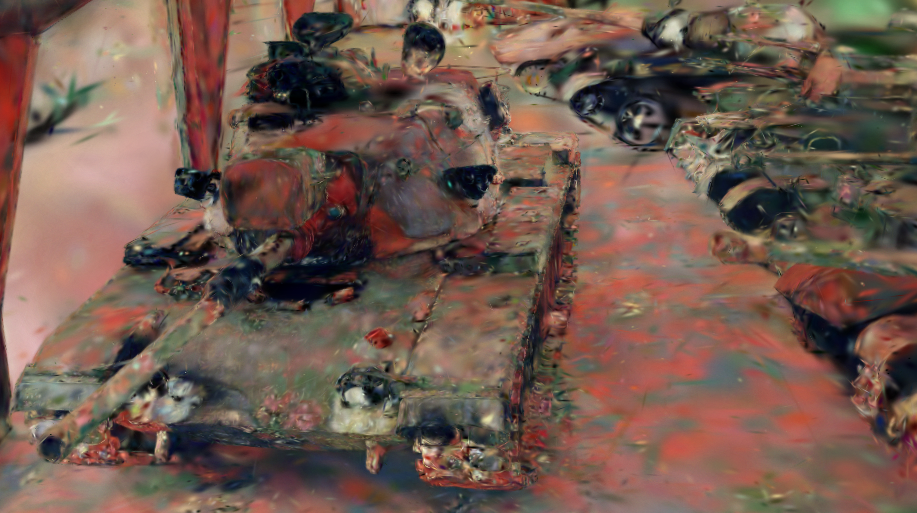} &
         \includegraphics[width=\imwidthc,height=\imheightc]{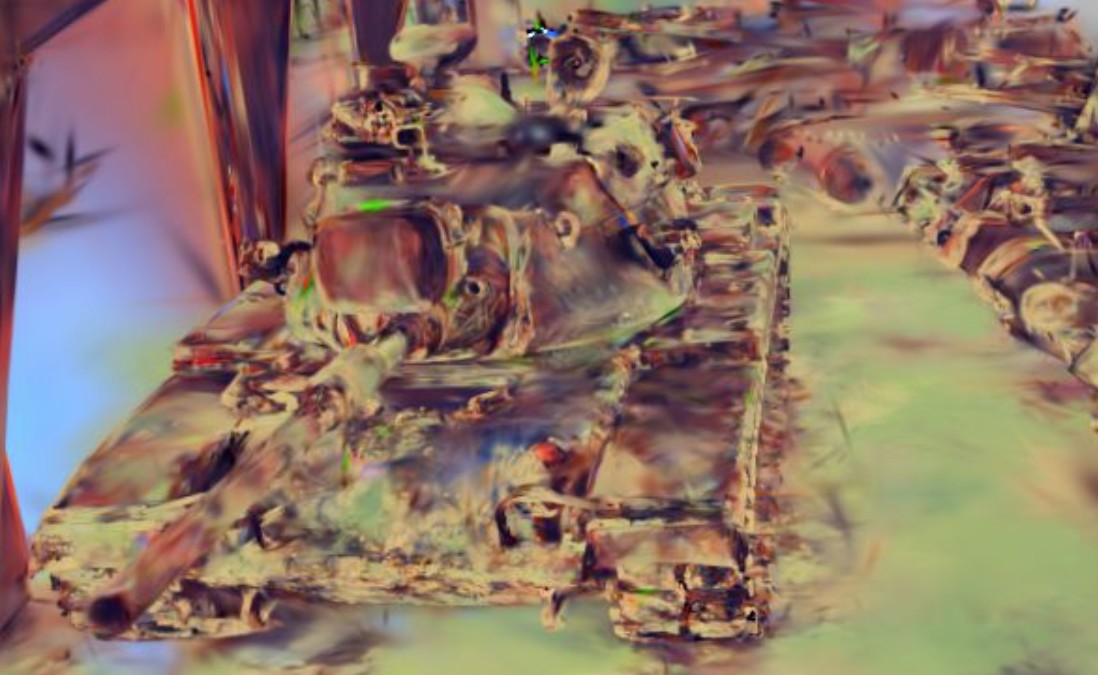} &
         \includegraphics[width=\imwidthc,height=\imheightc]{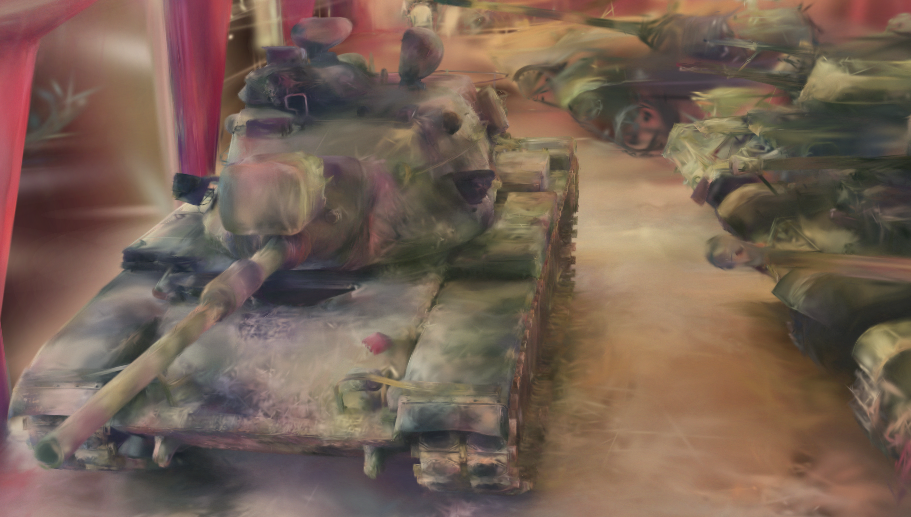} 
         \\

         & \multicolumn{1}{r}{LPIPS:} & 0.656 & 0.692 & 0.647
        \\

    \end{tabular}
    \caption{Comparison of our approach to other state-of-the-art techniques for different 3DGS scenes, views, and styles. LPIPS scores are reported between 10 views of the scene and calculated using \cite{zhang2018perceptual} to indicate consistency between views.}
    \label{fig:comparison-grid}
\end{figure*}

We compare our outputs with those of G-style \cite{kovacs2024} and StyleGaussian \cite{liu2024stylegaussian} 
since they are post reconstruction approaches like our method. 
Fig.~\ref{fig:comparison-grid} includes two different views of two different 3DGS scenes under style transfer. The outputs of G-style, Style Gaussian, and our approach are given. Additionally, LPIPS scores \cite{zhang2018perceptual} were computed for each scene on a 10-view set of the scene, indicating the consistency between views.

Overall, G-style has the best content and style preservation, likely due to the ability of G-style to move splats during the fine-tuning process. Our approach had the best consistency between views, but tends to have less sharp features in some areas. We note, however, that our approach has better color alignment than StyleGaussian in the examples that were tested. In general, our approach still provides reasonable quality outputs regardless of 3DGS size, but is better suited for single object stylization than for large scenes.


In addition to the rendered results, we compare timing information of our approach to other state-of-the-art 3DGS stylization methods. Because the nature of the stylizations differs, we separate the speeds into four speed categories: reconstruction, preprocessing, stylization, and rendering speeds. Reconstruction time includes the optimization process that generates the individual Gaussians given the input images of the scenes (some approaches include the stylization in this step). Preprocessing time includes any step needed to prepare the 3DGS scene for stylization (e.g. the graph construction process in our approach). Stylization time is the runtime of the incorporated stylization network or stylization process. Rendering time accounts for the time needed to generate a stylized image when viewing the scene.

To make these comparisons, we conducted our own timing experiments on an NVIDIA RTX 4090 desktop machine on the ``train'' 3DGS scene from Fig.~\ref{fig:style-grid}. The resulting speeds are shown in Table~\ref{tab:speed}. If an algorithm can be used on an already existing 3DGS, this was also noted as ``preoptimized'' in the table. As is shown, our method performs faster than existing approaches, with both preprocessing and stylization steps being completed in under a minute.

As an additional point, our approach is particularly advantageous for stylizing splats on lower-end hardware, since the graph construction process is implemented on the CPU with only the simple CNN-based stylization optionally running on the GPU rather than the CPU. This gives speeds around 1 minute even when no GPU is available. In comparison, fast methods like G-style \cite{kovacs2024} reported stylization speeds of 20-28 minutes for 360$^\circ$ scenes on an NVIDIA L4 GPU in Google Colab. Our approach is a viable option for stylization on standard consumer-based hardware.

\begin{table*}[t!]
    \centering
    \caption{Speed comparisons on the ``Train" example from Fig.~\ref{fig:style-grid}. Experiments were conducted on an NVIDIA RTX 4090 GPU with 24GB VRAM and an Intel i9 CPU. We also ran our method on an Mac M2 processor with no MPS acceleration to showcase the low hardware requirements for our stylization technique. *The SGSST time for this scene with similar hardware is self-reported in the original work \cite{galerne2024sgsst}. $^\dagger$The code for StyleGaussian is written to retrain the style transfer network for each individual 3DGS scene and these results are presented as such.}
    \begin{tabular}{|c|c|c|c|c|c|}
        \hline
        Method & Reconstruction & Preprocessing & Stylization & Rendering & Approx. Total (I/O included) \\
        \hline
         SGSST \cite{galerne2024sgsst} & *33 min & - & - & Real-time & *33 min \\
         G-style \cite{kovacs2024} & Preoptimized & 4.5 min & 8.7 min & Real-time & 12.5 min \\
         StyleGaussian \cite{liu2024stylegaussian} & Preoptimized & 4.08 min + $^\dagger$3hrs & 0.1 sec & Real-time & 4 min + $^\dagger$3hrs \\
         Ours & Preoptimized & 34.47 sec & 22.65 sec & Real-time & 1 min\\
         Ours (CPU Only) & Preoptimized & 34.22 sec & 56.65 sec & Real-time & 1.5 min \\
         Ours (Mac M2) & Preoptimized & 22.07 sec & 44.98 sec & Real-time & 1.25 min \\
         \hline
    \end{tabular}
    \label{tab:speed}
\end{table*}

\subsection{Ablation: Random Normals}

\begin{figure}[t]
\begin{center}
       \begin{tabular}{c}       
        \includegraphics[width=0.55\linewidth]{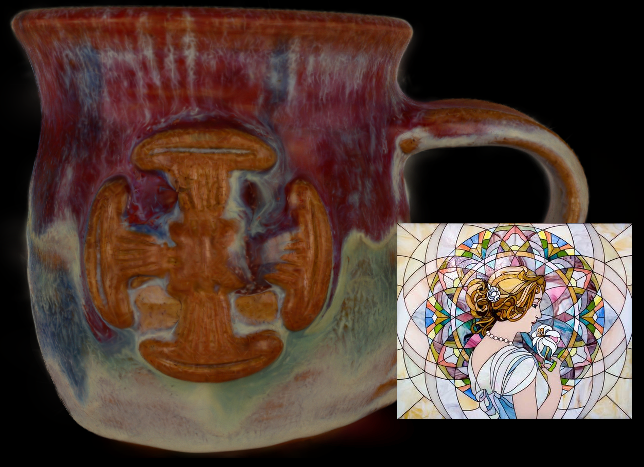}
        \\
        Original 3DGS and Style Image
        \\
        \includegraphics[width=0.55\linewidth]{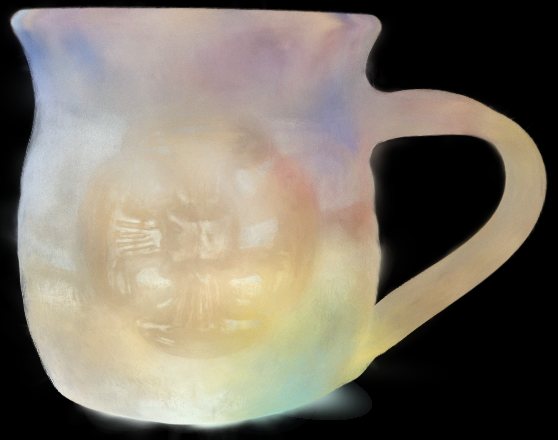}
        \\
        Style Transfer with Randomized Normals
        \\
        \includegraphics[width=0.55\linewidth]{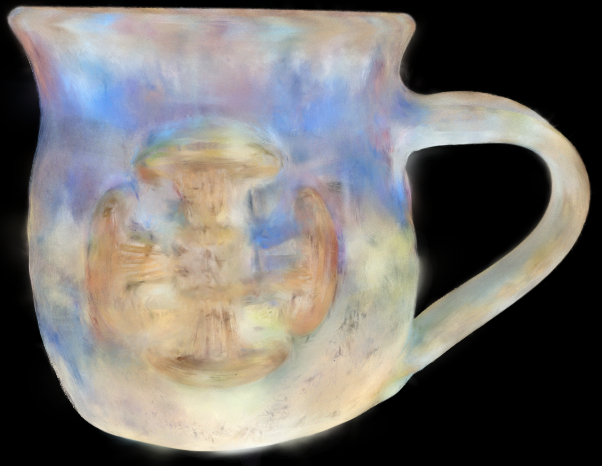}
        \\
        Style Transfer with Calculated Normals
    \end{tabular}
\end{center}
   \caption[Effects of Ball-Pivoting vs Random Normals on Style Transfer]{A 3D Gaussian splat of a simple object (Top). Accurately calculating normals is shown to be an important step of the style transfer since stylization with randomized normals (Middle) is excessively blurry compared to stylization with normals calculated with the Ball-Pivoting algorithm (Bottom).}
\label{fig:prettycupnoramls}
\end{figure}

Given that our technique operates on a pseudo implicit surface, it is worth investigating the need for accurate normals at all during the stylization. In even geometrically simple objects, however, the need for accurate normals is immediately apparent. Fig.~\ref{fig:prettycupnoramls} shows an example stylization of a 3DGS of a simple object. To highlight the impact of good accuracy, a style transfer with random normals was compared with a style transfer with Ball-Pivoting calculated normals. In the randomized normals example, the style transfer showed the characteristic blurriness of imperfect style transfer, likely due to the convolution operator not being able to preserve a global orientation. These kinds of artifacts are consistently present in all stylized 3DGS with randomized normals. The style transfer with calculated normals, meanwhile, greatly improves the stylization, with the decorative elements remaining distinct and different sections of the object showing distinct elements of the style.

\section{Conclusion}
The presented method is both novel and effective for performing stylization on 3D Gaussian splats. Additionally, its surface-based approach can run on preoptimized 3DGS scenes without ever rendering a view image and is fast even on standard consumer hardware.

Limitations of this approach include the inability to modify geometry during the stylization. Removing such a limitation could allow for stylized Gaussians to more closely match characteristic features of the style image. An additional limitation includes the reliance on a pseudo implicit surface. While we empirically found, in the 3DGS scenes we tested, that splats always tend to optimize near the surface of an object, more quantitative tests would need to be conducted to verify the universality of this claim. Additional considerations would need to be made for using our approach when a pseudo implicit surface is not present.

Future work could include changing the underlying stylization network. We used the CNN-based stylization approach of Li \etal \cite{li2019learning} because it was originally incorporated into Interpolated SelectionConv \cite{hart2023interpolated}, the surface-based method we build on. Improvements could occur by incorporating a modern vision transformer or diffusion-based style transfer network. Such models 
would require considerable modifications to the underlying graph structure and 
could be an avenue for future research. 

{
    \small
    \bibliographystyle{ieeenat_fullname}
    \bibliography{main}

@String(CVPR= {IEEE Conf. Comput. Vis. Pattern Recog.})

@String(ICCV= {Int. Conf. Comput. Vis.})

@String(TOG= {ACM Trans. Graph.})

@String(ICASSP=	{ICASSP})

@String(CVPR  = {CVPR})

@String(ICCV  = {ICCV})

@String(TOG   = {ACM TOG})

@article{mildenhall2021nerf,
  title={Nerf: Representing scenes as neural radiance fields for view synthesis},
  author={Mildenhall, Ben and Srinivasan, Pratul P and Tancik, Matthew and Barron, Jonathan T and Ramamoorthi, Ravi and Ng, Ren},
  journal={Communications of the ACM},
  volume={65},
  number={1},
  pages={99--106},
  year={2021},
  publisher={ACM New York, NY, USA}
}

@article{schwarz2022voxgraf,
  title={Voxgraf: Fast 3d-aware image synthesis with sparse voxel grids},
  author={Schwarz, Katja and Sauer, Axel and Niemeyer, Michael and Liao, Yiyi and Geiger, Andreas},
  journal={Advances in Neural Information Processing Systems},
  volume={35},
  pages={33999--34011},
  year={2022}
}

@article{kerbl20233d,
  title={3d gaussian splatting for real-time radiance field rendering.},
  author={Kerbl, Bernhard and Kopanas, Georgios and Leimk{\"u}hler, Thomas and Drettakis, George},
  journal={ACM Trans. Graph.},
  volume={42},
  number={4},
  pages={139--1},
  year={2023}
}

@article{jain2024stylesplat,
  title={Stylesplat: 3d object style transfer with gaussian splatting},
  author={Jain, Sahil and Kuthiala, Avik and Sethi, Prabhdeep Singh and Saxena, Prakanshul},
  journal={arXiv preprint arXiv:2407.09473},
  year={2024}
}

@article{xu2024instantmesh,
  title={Instantmesh: Efficient 3d mesh generation from a single image with sparse-view large reconstruction models},
  author={Xu, Jiale and Cheng, Weihao and Gao, Yiming and Wang, Xintao and Gao, Shenghua and Shan, Ying},
  journal={arXiv preprint arXiv:2404.07191},
  year={2024}
}

@article{sitzmann2019scene,
  title={Scene representation networks: Continuous 3d-structure-aware neural scene representations},
  author={Sitzmann, Vincent and Zollh{\"o}fer, Michael and Wetzstein, Gordon},
  journal={Advances in Neural Information Processing Systems},
  volume={32},
  year={2019}
}

@inproceedings{sitzmann2019deepvoxels,
  title={Deepvoxels: Learning persistent 3d feature embeddings},
  author={Sitzmann, Vincent and Thies, Justus and Heide, Felix and Nie{\ss}ner, Matthias and Wetzstein, Gordon and Zollhofer, Michael},
  booktitle={Proceedings of the IEEE/CVF Conference on Computer Vision and Pattern Recognition},
  pages={2437--2446},
  year={2019}
}

@article{hedman2018deep,
  title={Deep blending for free-viewpoint image-based rendering},
  author={Hedman, Peter and Philip, Julien and Price, True and Frahm, Jan-Michael and Drettakis, George and Brostow, Gabriel},
  journal={ACM Transactions on Graphics (ToG)},
  volume={37},
  number={6},
  pages={1--15},
  year={2018},
  publisher={ACM New York, NY, USA}
}

@incollection{chen2023view,
  title={View interpolation for image synthesis},
  author={Chen, Shenchang Eric and Williams, Lance},
  booktitle={Seminal Graphics Papers: Pushing the Boundaries, Volume 2},
  publisher={Association for Computing Machinery},
  pages={423--432},
  year={1993}
}

@article{saroha2024gaussian,
  title={Gaussian splatting in style},
  author={Saroha, Abhishek and Gladkova, Mariia and Curreli, Cecilia and Muhle, Dominik and Yenamandra, Tarun and Cremers, Daniel},
  journal={arXiv preprint arXiv:2403.08498},
  year={2024}
}

@inproceedings{hu2024low,
  title={Low Latency Point Cloud Rendering with Learned Splatting},
  author={Hu, Yueyu and Gong, Ran and Sun, Qi and Wang, Yao},
  booktitle={Proceedings of the IEEE/CVF Conference on Computer Vision and Pattern Recognition},
  pages={5752--5761},
  year={2024}
}

@inproceedings{kovacs2024,
  title={G-Style: Stylized Gaussian Splatting},
  author={Kov{\'a}cs, {\'A}ron Samuel and Hermosilla, Pedro and Raidou, Renata G},
  booktitle={Computer Graphics Forum},
  volume={43},
  number={7},
  pages={e15259},
  year={2024},
  organization={Wiley Online Library}
}

@inproceedings{gatys2016image,
  title={Image style transfer using convolutional neural networks},
  author={Gatys, Leon A and Ecker, Alexander S and Bethge, Matthias},
  booktitle={Proceedings of the IEEE conference on computer vision and pattern recognition},
  pages={2414--2423},
  year={2016}
}

@article{simonyan2014very,
  title={Very deep convolutional networks for large-scale image recognition},
  author={Simonyan, Karen},
  journal={arXiv preprint arXiv:1409.1556},
  year={2014}
}

@inproceedings{zhang2022arf,
  title={Arf: Artistic radiance fields},
  author={Zhang, Kai and Kolkin, Nick and Bi, Sai and Luan, Fujun and Xu, Zexiang and Shechtman, Eli and Snavely, Noah},
  booktitle={European Conference on Computer Vision},
  pages={717--733},
  year={2022},
  organization={Springer}
}

@inproceedings{johnson2016perceptual,
  title={Perceptual losses for real-time style transfer and super-resolution},
  author={Johnson, Justin and Alahi, Alexandre and Fei-Fei, Li},
  booktitle={Computer Vision--ECCV 2016: 14th European Conference, Amsterdam, The Netherlands, October 11-14, 2016, Proceedings, Part II 14},
  pages={694--711},
  year={2016},
  organization={Springer}
}

@article{galerne2024sgsst,
  title={SGSST: Scaling Gaussian Splatting StyleTransfer},
  author={Galerne, Bruno and Wang, Jianling and Raad, Lara and Morel, Jean-Michel},
  journal={arXiv preprint arXiv:2412.03371},
  year={2024}
}

@inproceedings{kovacs2024surface,
  title={Surface-aware Mesh Texture Synthesis with Pre-trained 2D CNNs},
  author={Kov{\'a}cs, {\'A}ron Samuel and Hermosilla, Pedro and Raidou, Renata G},
  booktitle={Computer Graphics Forum},
  volume={43},
  number={2},
  pages={e15016},
  year={2024},
  organization={Wiley Online Library}
}

@inproceedings{hart2023interpolated,
  title={Interpolated SelectionConv for spherical images and surfaces},
  author={Hart, David and Whitney, Michael and Morse, Bryan},
  booktitle={Proceedings of the IEEE/CVF Winter Conference on Applications of Computer Vision},
  pages={321--330},
  year={2023}
}

@inproceedings{li2019learning,
  title={Learning linear transformations for fast image and video style transfer},
  author={Li, Xueting and Liu, Sifei and Kautz, Jan and Yang, Ming-Hsuan},
  booktitle={Proceedings of the IEEE/CVF conference on computer vision and pattern recognition},
  pages={3809--3817},
  year={2019}
}

@inproceedings{huang2017arbitrary,
  title={Arbitrary style transfer in real-time with adaptive instance normalization},
  author={Huang, Xun and Belongie, Serge},
  booktitle={Proceedings of the IEEE international conference on computer vision},
  pages={1501--1510},
  year={2017}
}

@incollection{liu2024stylegaussian,
  title={Stylegaussian: Instant 3d style transfer with gaussian splatting},
  author={Liu, Kunhao and Zhan, Fangneng and Xu, Muyu and Theobalt, Christian and Shao, Ling and Lu, Shijian},
  booktitle={SIGGRAPH Asia 2024 Technical Communications},
  pages={1--4},
  year={2024},
  publisher={SIGGRAPH}
}

@article{jaganathan2024ice,
  title={ICE-G: Image Conditional Editing of 3D Gaussian Splats},
  author={Jaganathan, Vishnu and Huang, Hannah Hanyun and Irshad, Muhammad Zubair and Jampani, Varun and Raj, Amit and Kira, Zsolt},
  journal={arXiv preprint arXiv:2406.08488},
  year={2024}
}

@article{zhang2024stylizedgs,
  title={Stylizedgs: Controllable stylization for 3d gaussian splatting},
  author={Zhang, Dingxi and Yuan, Yu-Jie and Chen, Zhuoxun and Zhang, Fang-Lue and He, Zhenliang and Shan, Shiguang and Gao, Lin},
  journal={arXiv preprint arXiv:2404.05220},
  year={2024}
}

@inproceedings{hart2022selectionconv,
  title={SelectionConv: convolutional neural networks for non-rectilinear image data},
  author={Hart, David and Whitney, Michael and Morse, Bryan},
  booktitle={European Conference on Computer Vision},
  pages={317--333},
  year={2022},
  organization={Springer}
}

@article{bernardini2002ball,
  title={The ball-pivoting algorithm for surface reconstruction},
  author={Bernardini, Fausto and Mittleman, Joshua and Rushmeier, Holly and Silva, Cl{\'a}udio and Taubin, Gabriel},
  journal={IEEE transactions on visualization and computer graphics},
  volume={5},
  number={4},
  pages={349--359},
  year={2002},
  publisher={IEEE}
}

@InProceedings{styleID,
    author    = {Chung, Jiwoo and Hyun, Sangeek and Heo, Jae-Pil},
    title     = {Style Injection in Diffusion: A Training-free Approach for Adapting Large-scale Diffusion Models for Style Transfer},
    booktitle = {Proceedings of the IEEE/CVF Conference on Computer Vision and Pattern Recognition (CVPR)},
    month     = {June},
    year      = {2024},
    pages     = {8795-8805}
}

@inproceedings{deng2022stytr2,
  title={Stytr2: Image style transfer with transformers},
  author={Deng, Yingying and Tang, Fan and Dong, Weiming and Ma, Chongyang and Pan, Xingjia and Wang, Lei and Xu, Changsheng},
  booktitle={Proceedings of the IEEE/CVF conference on computer vision and pattern recognition},
  pages={11326--11336},
  year={2022}
}

@article{li2017universal,
  title={Universal style transfer via feature transforms},
  author={Li, Yijun and Fang, Chen and Yang, Jimei and Wang, Zhaowen and Lu, Xin and Yang, Ming-Hsuan},
  journal={Advances in neural information processing systems},
  volume={30},
  year={2017}
}

@InProceedings{niedermayr2024compress1,
    author    = {Niedermayr, Simon and Stumpfegger, Josef and Westermann, R\"udiger},
    title     = {Compressed 3D Gaussian Splatting for Accelerated Novel View Synthesis},
    booktitle = {Proceedings of the IEEE/CVF Conference on Computer Vision and Pattern Recognition (CVPR)},
    month     = {June},
    year      = {2024},
    pages     = {10349-10358}
}

@misc{fan2023compress2,
  author    = {Zhiwen Fan and Kevin Wang and Kairun Wen and Zehao Zhu and Dejia Xu and Zhangyang Wang},
  title     = {LightGaussian: Unbounded 3D Gaussian Compression with 15x Reduction and 200+ FPS},
  year      ={2023}, 
  eprint    ={2311.17245}, 
  archivePrefix={arXiv}, 
  primaryClass={cs.CV},
}

@article{navaneet2023compress3,
          title={Compact3D: Smaller and Faster Gaussian Splatting with Vector Quantization},
          author={Navaneet, KL and Meibodi, Kossar Pourahmadi and Koohpayegani, Soroush Abbasi and Pirsiavash, Hamed},
          journal={arXiv preprint arXiv:2311.18159},
          year={2023}
        }

@InProceedings{lee2024compress4,
  author    = {Lee, Joo Chan and Rho, Daniel and Sun, Xiangyu and Ko, Jong Hwan and Park, Eunbyung},
  title     = {Compact 3D Gaussian Representation for Radiance Field},
  booktitle = {Proceedings of the IEEE/CVF Conference on Computer Vision and Pattern Recognition (CVPR)},
  year      = {2024},
  pages     = {21719-21728}
}

@article{morgenstern2023compress5,
  title={Compact 3D Scene Representation via Self-Organizing Gaussian Grids},
  author={Morgenstern, Wieland and Barthel, Florian and Hilsmann, Anna and Eisert, Peter},
  journal={arXiv preprint arXiv:2312.13299},
  year={2023}
}

@misc{igs2gs,
         author = {Vachha, Cyrus and Haque, Ayaan},
         title = {Instruct-GS2GS: Editing 3D Gaussian Splats with Instructions},
         year = {2024},
         url = {https://instruct-gs2gs.github.io/}
        }

@misc{gsedit,
      title={GSEdit: Efficient Text-Guided Editing of 3D Objects via Gaussian Splatting}, 
      author={Francesco Palandra and Andrea Sanchietti and Daniele Baieri and Emanuele Rodolà},
      year={2024},
      eprint={2403.05154},
      archivePrefix={arXiv},
      primaryClass={cs.CV},
      url={https://arxiv.org/abs/2403.05154}, 
}

@inproceedings{gatys2017controlling,
  title={Controlling perceptual factors in neural style transfer},
  author={Gatys, Leon A and Ecker, Alexander S and Bethge, Matthias and Hertzmann, Aaron and Shechtman, Eli},
  booktitle={Proceedings of the IEEE conference on computer vision and pattern recognition},
  pages={3985--3993},
  year={2017}
}

@INPROCEEDINGS{lightfield1,
  author={Hart, David and Greenland, Jessica and Morse, Bryan},
  booktitle={2020 IEEE Winter Conference on Applications of Computer Vision (WACV)}, 
  title={Style Transfer for Light Field Photography}, 
  year={2020},
  volume={},
  number={},
  pages={99-108},
  doi={10.1109/WACV45572.2020.9093478}}

@INPROCEEDINGS{lightfield2,
  author={Egan, Dónal and Alain, Martin and Smolic, Aljosa},
  booktitle={ICASSP 2021 - 2021 IEEE International Conference on Acoustics, Speech and Signal Processing (ICASSP)}, 
  title={Light Field Style Transfer with Local Angular Consistency}, 
  year={2021},
  volume={},
  number={},
  pages={2300-2304},
  doi={10.1109/ICASSP39728.2021.9414689}}

@INPROCEEDINGS{huang2017,
  author={Huang, Haozhi and Wang, Hao and Luo, Wenhan and Ma, Lin and Jiang, Wenhao and Zhu, Xiaolong and Li, Zhifeng and Liu, Wei},
  booktitle={2017 IEEE Conference on Computer Vision and Pattern Recognition (CVPR)}, 
  title={Real-Time Neural Style Transfer for Videos}, 
  year={2017},
  volume={},
  number={},
  pages={7044-7052},
  doi={10.1109/CVPR.2017.745}}

@InProceedings{Ye_2025_CVPR,
    author    = {Ye, Zixuan and Huang, Huijuan and Wang, Xintao and Wan, Pengfei and Zhang, Di and Luo, Wenhan},
    title     = {StyleMaster: Stylize Your Video with Artistic Generation and Translation},
    booktitle = {Proceedings of the Computer Vision and Pattern Recognition Conference (CVPR)},
    month     = {June},
    year      = {2025},
    pages     = {2630-2640}
}

@article{ruder2018artistic,
  title={Artistic style transfer for videos and spherical images},
  author={Ruder, Manuel and Dosovitskiy, Alexey and Brox, Thomas},
  journal={International Journal of Computer Vision},
  volume={126},
  number={11},
  pages={1199--1219},
  year={2018},
  publisher={Springer}
}

@article{ebsynth,
author = {Jamri\v{s}ka, Ond\v{r}ej and Sochorov\'{a}, \v{S}\'{a}rka and Texler, Ond\v{r}ej and Luk\'{a}\v{c}, Michal and Fi\v{s}er, Jakub and Lu, Jingwan and Shechtman, Eli and S\'{y}kora, Daniel},
title = {Stylizing video by example},
year = {2019},
issue_date = {August 2019},
publisher = {Association for Computing Machinery},
address = {New York, NY, USA},
volume = {38},
number = {4},
issn = {0730-0301},
url = {https://doi.org/10.1145/3306346.3323006},
doi = {10.1145/3306346.3323006},
journal = {ACM Trans. Graph.},
month = jul,
articleno = {107},
numpages = {11}
}

@article{stylit,
author = {Fi\v{s}er, Jakub and Jamri\v{s}ka, Ond\v{r}ej and Luk\'{a}\v{c}, Michal and Shechtman, Eli and Asente, Paul and Lu, Jingwan and S\'{y}kora, Daniel},
title = {StyLit: illumination-guided example-based stylization of 3D renderings},
year = {2016},
issue_date = {July 2016},
publisher = {Association for Computing Machinery},
address = {New York, NY, USA},
volume = {35},
number = {4},
issn = {0730-0301},
url = {https://doi.org/10.1145/2897824.2925948},
doi = {10.1145/2897824.2925948},
journal = {ACM Trans. Graph.},
month = jul,
articleno = {92},
numpages = {11}
}

@Article{styblit,
  author =  "Daniel S\'{y}kora and Ond\v{r}ej Jamri\v{s}ka and Ond\v{r}ej Texler 
             and Jakub Fi\v{s}er and Michal Luk\'{a}\v{c} and Jingwan Lu and Eli Shechtman",
  title =   "{StyleBlit}: Fast Example-Based Stylization with Local Guidance",
  journal = "Computer Graphics Forum",
  volume =  38,
  number =  2,
  pages =   "83--91",
  year =    2019,
}

@article{fluidstyle,
author = {Kim, Byungsoo and Azevedo, Vinicius C. and Gross, Markus and Solenthaler, Barbara},
title = {Lagrangian neural style transfer for fluids},
year = {2020},
issue_date = {August 2020},
publisher = {Association for Computing Machinery},
address = {New York, NY, USA},
volume = {39},
number = {4},
issn = {0730-0301},
url = {https://doi.org/10.1145/3386569.3392473},
doi = {10.1145/3386569.3392473},
journal = {ACM Trans. Graph.},
month = aug,
articleno = {52},
numpages = {10}
}

@article{Open3D,
    author    = {Qian-Yi Zhou and Jaesik Park and Vladlen Koltun},
    title     = {{Open3D}: {A} Modern Library for {3D} Data Processing},
    journal   = {arXiv:1801.09847},
    year      = {2018},
}

@article{Knapitsch2017,
    author    = {Arno Knapitsch and Jaesik Park and Qian-Yi Zhou and Vladlen Koltun},
    title     = {Tanks and Temples: Benchmarking Large-Scale Scene Reconstruction},
    journal   = {ACM Transactions on Graphics},
    volume    = {36},
    number    = {4},
    year      = {2017},
}

@misc{kobranov2024splats,
  author       = {Vlad Kobranov},
  title        = {splats},
  year         = {2024},
  howpublished = {\url{https://huggingface.co/VladKobranov/splats}},
  note         = {Accessed: 2025-07-15}
}

@misc{Scaniverse,
  author = {{Niantic}},
  title = {{Scaniverse: 3D Scanner}},
  howpublished = {Mobile App},
  year = {2021},
  note = {Available at: \url{https://apps.apple.com/us/app/scaniverse-3d-scanner/id1541433223} and \url{https://play.google.com/store/apps/details?id=com.nianticlabs.scaniverse}},
  url = {https://scaniverse.com/}
}

@InProceedings{Zhang_2023_inst,
 author    = {Zhang, Yuxin and Huang, Nisha and Tang, Fan and Huang, Haibin and Ma, Chongyang and Dong, Weiming and Xu, Changsheng},
 title     = {Inversion-Based Style Transfer With Diffusion Models},
 booktitle = {Proceedings of the IEEE/CVF Conference on Computer Vision and Pattern Recognition (CVPR)},
 month     = {June},
 year      = {2023},
 pages     = {10146-10156}
}

@INPROCEEDINGS{wu2021styleformer,
  author={Wu, Xiaolei and Hu, Zhihao and Sheng, Lu and Xu, Dong},
  booktitle={2021 IEEE/CVF International Conference on Computer Vision (ICCV)}, 
  title={StyleFormer: Real-time Arbitrary Style Transfer via Parametric Style Composition}, 
  year={2021},
  volume={},
  number={},
  pages={14598-14607},
  doi={10.1109/ICCV48922.2021.01435}}

@inproceedings{painterlyrendering,
author = {Hertzmann, Aaron},
title = {Painterly rendering with curved brush strokes of multiple sizes},
year = {1998},
isbn = {0897919998},
publisher = {Association for Computing Machinery},
address = {New York, NY, USA},
url = {https://doi.org/10.1145/280814.280951},
doi = {10.1145/280814.280951},
booktitle = {Proceedings of the 25th Annual Conference on Computer Graphics and Interactive Techniques},
pages = {453–460},
numpages = {8},
series = {SIGGRAPH '98}
}

@article{liu2023stylerf,
  author = {Kunhao Liu and Fangneng Zhan and Yiwen Chen and Jiahui Zhang and Yingchen Yu and Abdulmotaleb El Saddik and Shijian Lu and Eric Xing},
  title = {StyleRF: Zero-shot 3D Style Transfer of Neural Radiance Fields},
  booktitle = {Proc. IEEE Conf. on Computer Vision and Pattern Recognition (CVPR)},
  year = {2023},
}

@inproceedings{chiang2022stylizing,
  title={Stylizing 3d scene via implicit representation and hypernetwork},
  author={Chiang, Pei-Ze and Tsai, Meng-Shiun and Tseng, Hung-Yu and Lai, Wei-Sheng and Chiu, Wei-Chen},
  booktitle={Proceedings of the IEEE/CVF winter conference on applications of computer vision},
  pages={1475--1484},
  year={2022}
}

@InProceedings{Yu_2024_CVPR,
    author    = {Yu, Zehao and Chen, Anpei and Huang, Binbin and Sattler, Torsten and Geiger, Andreas},
    title     = {Mip-Splatting: Alias-free 3D Gaussian Splatting},
    booktitle = {Proceedings of the IEEE/CVF Conference on Computer Vision and Pattern Recognition (CVPR)},
    month     = {June},
    year      = {2024},
    pages     = {19447-19456}
}

@InProceedings{Fu_2024_CVPR,
    author    = {Fu, Yang and Liu, Sifei and Kulkarni, Amey and Kautz, Jan and Efros, Alexei A. and Wang, Xiaolong},
    title     = {COLMAP-Free 3D Gaussian Splatting},
    booktitle = {Proceedings of the IEEE/CVF Conference on Computer Vision and Pattern Recognition (CVPR)},
    month     = {June},
    year      = {2024},
    pages     = {20796-20805}
}

@inproceedings{Huang2024gaussian2D,
author = {Huang, Binbin and Yu, Zehao and Chen, Anpei and Geiger, Andreas and Gao, Shenghua},
title = {2D Gaussian Splatting for Geometrically Accurate Radiance Fields},
year = {2024},
isbn = {9798400705250},
publisher = {Association for Computing Machinery},
address = {New York, NY, USA},
url = {https://doi.org/10.1145/3641519.3657428},
doi = {10.1145/3641519.3657428},
booktitle = {ACM SIGGRAPH 2024 Conference Papers},
articleno = {32},
numpages = {11},
location = {Denver, CO, USA},
series = {SIGGRAPH '24}
}

@inproceedings{zhang2018perceptual,
  title={The Unreasonable Effectiveness of Deep Features as a Perceptual Metric},
  author={Zhang, Richard and Isola, Phillip and Efros, Alexei A and Shechtman, Eli and Wang, Oliver},
  booktitle={CVPR},
  year={2018}
}
}

\end{document}


\title{Optimization-Free Style Transfer for 3D Gaussian Splats - Supplemental Materials}

\maketitle

\section{Point Filtering}

In a practical sense, individual Gaussian splats tend to lie along implicit geometric surfaces, or at the very least tend to cluster rather than occupying isolated points in space. In most situations, any Gaussian splats that are highly isolated are likely to be ``noisy" data points or artifacts left over from the training steps when the representation was first generated from image data.  To reduce this noise, a filtering mechanism is implemented as preprocessing step to our graph construction pipeline. 
In addition to being visually unappealing, these noisy data potentially distort or even obscure content and reduce the accuracy of style transfers. 
With the assumption that most geometric objects tend to be contiguous and will be represented by splats that roughly follow along an implicit surface, we conclude that the more isolated a point is, and the more distant it is from any implicit surface, the more likely it is to be a noisy data point and a candidate for filtering out. 
Handily, both isolation from neighboring points and distance from an implicit surface can be calculated at the same time in our method. 

The filtering method is as follows: the Gaussian splat is converted to a sparse point cloud comprised of the central points of each individual Gaussian splat, which we define as set $P$. 
For each $p \in P$, the KNN algorithm is used to determine the neighborhood $\mathcal{N}$ about that point. In that neighborhood, the midpoint $p_m$ is calculated as the mean of all points in the neighborhood. The distance from the neighborhood average is then simply $p-p_m$:

\[\ d = (p - p_m) \text{ where } p_m = \frac{1}{K}\sum_{j \in \mathcal{N}} p_j \]

\begin{figure}
    \centering
    \includegraphics[width=0.75\linewidth]{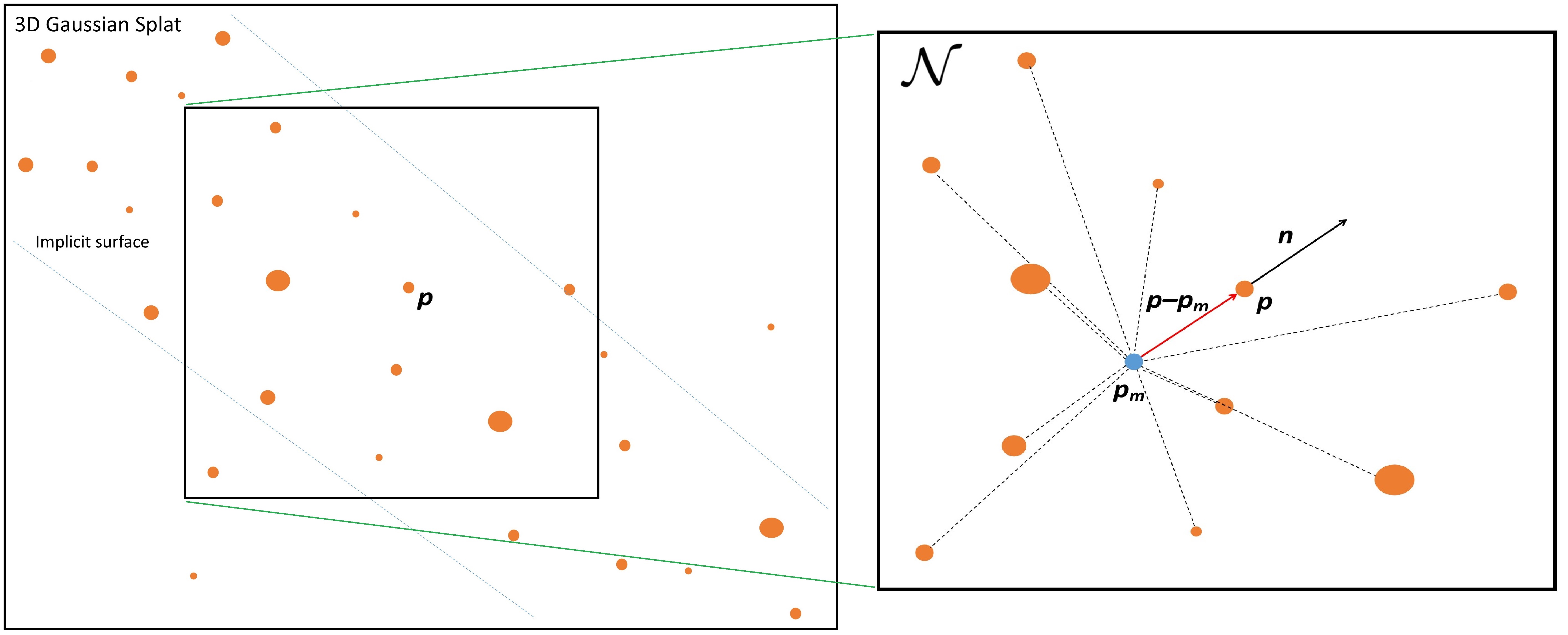}
    \caption{Within the point cloud derived from a 3D Gaussian splat, points in a neighborhood about some selected point $p$ are used to determine the mean point $p_m$ from which the distance $d$ is derived. Insofar as $p$ is on or within the implicit surface, $d$ should be small. The value of $d$ increases when $p$ is at a greater distance from other points in its neighborhood, i.e when $p$ does not lie in or on the implicit surface. }
    \label{fig:normalneighborhood}
\end{figure}

\noindent Figure \ref{fig:normalneighborhood} provides a visual demonstration of this mathematical operation. 

\begin{figure}
    \centering
    \includegraphics[width=0.75\linewidth]{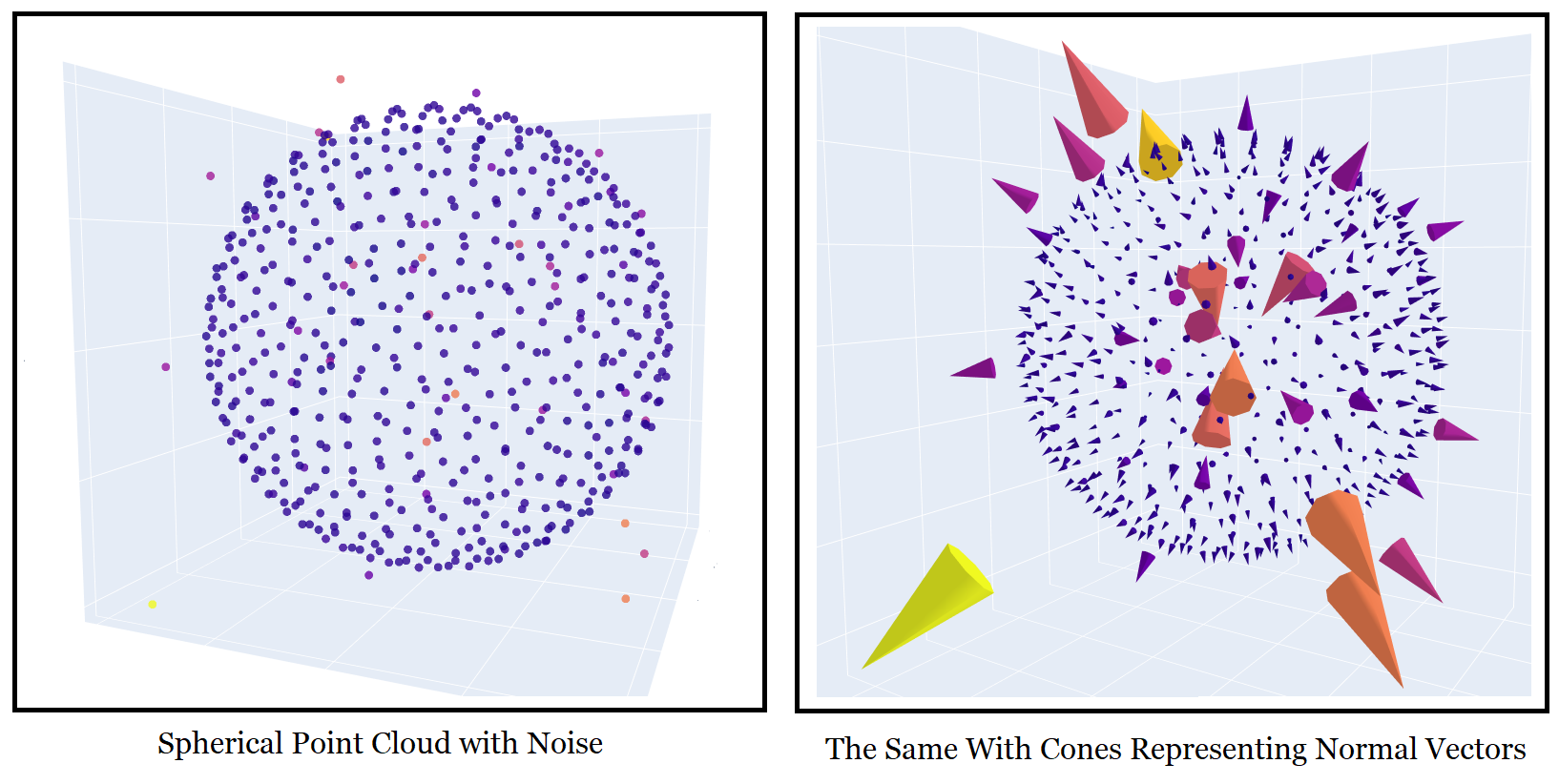}
    \caption{A spherical point cloud with a large number of points added at random distances from the surface to simulate noise (Left) and the same point cloud with cones representing the vectors between $p_m$ and $p$ (Right) demonstrating how the distance between $p_m$ and $p$ strongly segregates points close to the surface from those far from the surface.}
    \label{fig:normalcones}
\end{figure}

The calculated distance $d$ for isolated points is much greater than the $d$ of points in or on the implicit surface or for points that are in tightly clustered groups. Figure \ref{fig:normalcones} visualizes this phenomenon with a toy example. The filtering process then simply filters out the points with the top distances according to a user defined percentile. Filtering those points reduces the noise in the 3D Gaussian splat and better represents the implicit surface.



This filtering showed a high degree of reliability in segregating between individual Gaussian splats centered on points on and along the surface of objects being represented and those with large offsets from that surface. Some well-trained 3D Gaussian splats from public repositories required no filtering. Other publicly available 3DGSs, and in particular user-generated Gaussian splats, do require filtering. User-generated splats suffered from low training times and low-resolution reference images, and thus contained a high percentage of artifacts and noise. 

To study the efficacy of our filtering approach, we examined two different cases. The first was a test case constructed by generating a geometrically simple Gaussian splat containing noisy points. The second case involved a user-generated Gaussian splats.

\begin{figure}
    \centering
    \includegraphics[width=0.80\linewidth]{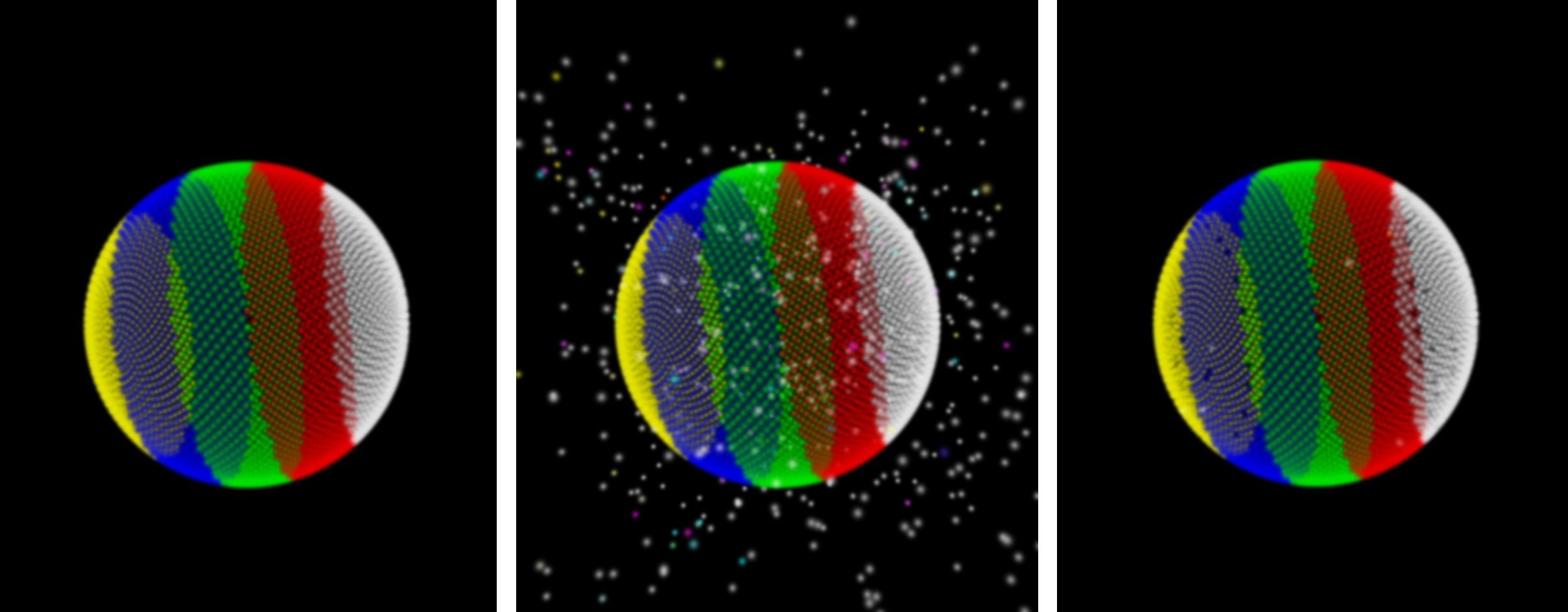}
    \caption[3DGS Sphere, with Noise, Before and After Filtering]{A 3D Gaussian of a sphere (Left) with a large number of points added at random distances from the surface to simulate noise (Middle) and the same after our filtering approach (Right)}
    \label{fig:spherefiltered}
\end{figure}

For the first case, a programmatically-derived Gaussian splat made up of 5000 equal sized Gaussians distributed at equidistant points along the surface of a unit sphere was generated. This was seeded with an additional 500 Gaussian splats centered on points offset from the surface. Offsetting was accomplished by randomly dispersing the points within a 2x2x2 volume centered on the unit sphere. Our filtering approach was applied to remove points not along the surface of the sphere. This obtained a filtering accuracy of 98.87\%. Results are shown in Figure \ref{fig:spherefiltered}.

\begin{figure}[t]
\begin{center}
       \begin{tabular}{cc}
        \includegraphics[width=0.40\linewidth]{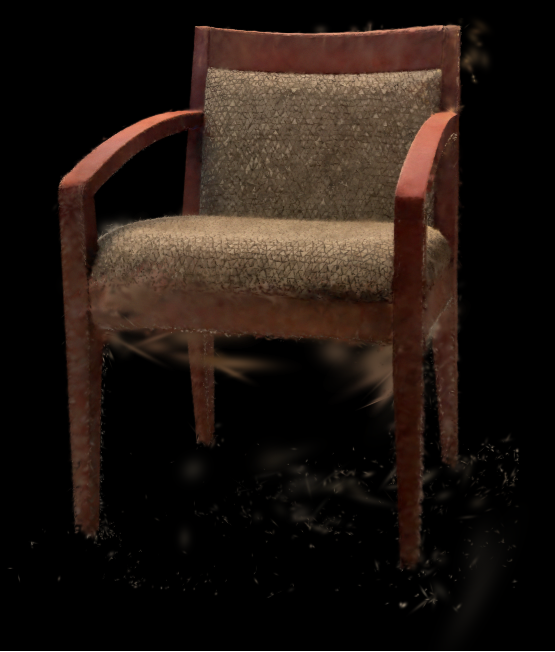}
&
        \includegraphics[width=0.40\linewidth]{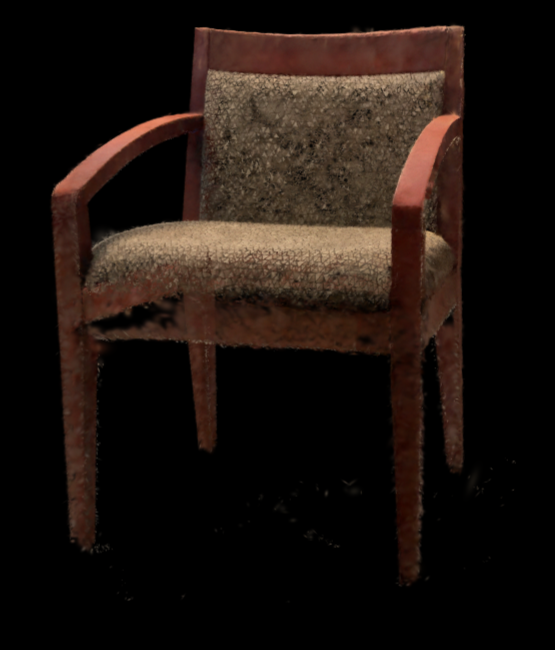}
        \\
        Original 3DGS
&
        Filtered 3DGS
        \\
        \includegraphics[width=0.40\linewidth]{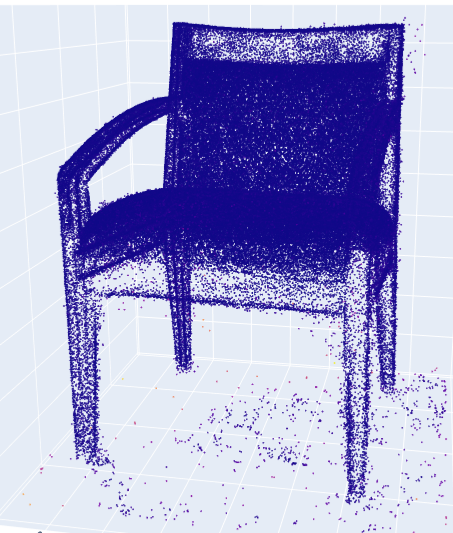}
&
        \includegraphics[width=0.40\linewidth]{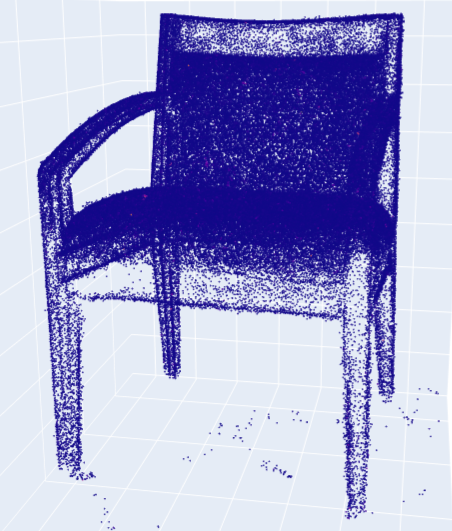}
        \\
        Original 3DGS as a Point Cloud 
&
        Filtered 3DGS as a Point Cloud
    \end{tabular}
\end{center}
   \caption{A 3D Gaussian splat that contains a large number of artifacts (Top) and noise both before (Left) and after (Right) being filtered. Simplified representations of the 3DGS formed by creating a point cloud with the central point of each individual Gaussian splat (Bottom) demonstrate how isolated points and points well-outside the implicit surface are filtered.}
\label{fig_chairfiltered}
\end{figure}

In the second case, the Scaniverse mobile app was used to create a 3D Gaussian splat of a household object, in this case a chair. This file was found to be particularly noisy and so was an excellent candidate for this study. Before and after filtering results are presented in Figure \ref{fig_chairfiltered}. 
As is shown, isolated points are preferentially filtered, while points along the implicit surface of the object, in this case a chair, are not filtered. One limitation of this approach, however, is the inability to filter points that are noisy but tightly clustered.

